%% file: main.tex
\documentclass[10pt,twocolumn,letterpaper]{article}

\usepackage{cvpr}              % To produce the CAMERA-READY version
\usepackage{xcolor} % For color
\usepackage{amssymb} % For symbols
\usepackage{style}
\usepackage[bottom]{footmisc}
\usepackage{tcolorbox}
\definecolor{promptbg}{RGB}{240, 240, 240}
\definecolor{promptborder}{RGB}{200, 200, 200}

\usepackage{transparent}
\newcommand\blfootnote[1]{%
  \begingroup
  \renewcommand\thefootnote{}\footnote{#1}%
  \addtocounter{footnote}{-1}%
  \endgroup
}
\input{preamble}

\newcommand\SupplementaryMaterials{%
  \xdef\presupfigures{\arabic{figure}}% save the current figure number
  \xdef\presuptables{\arabic{table}}% save the current figure number
  \xdef\presupsections{\arabic{section}}% save the current section number
  \renewcommand\thefigure{S\fpeval{\arabic{figure}-\presupfigures}}
  \renewcommand\thetable{S\fpeval{\arabic{table}-\presuptables}}
  \renewcommand\thesection{S\fpeval{\arabic{section}-\presupsections}}
}

\newcommand{\methodname}{{{AIpparel}}\xspace}
\definecolor{cvprblue}{rgb}{0.21,0.49,0.74}
\usepackage[pagebackref,breaklinks,colorlinks,citecolor=cvprblue,urlcolor=cvprblue,linkcolor=cvprblue]{hyperref}

\newcommand{\cmark}{\textcolor{green}{\checkmark}} % Green checkmark
\newcommand{\xmark}{\textcolor{red}{\ding{55}}} % Red cross
\newcommand{\changelocaltocdepth}[1]{%
  \addtocontents{toc}{\protect\setcounter{tocdepth}{#1}}%
  \setcounter{tocdepth}{#1}%
}

\def\paperID{3945} % *** Enter the Paper ID here
\def\confName{CVPR}
\def\confYear{2025}

\vspace{-0.5cm}
\title{\methodname: A Multimodal Foundation Model for Digital Garments\vspace{-0.5cm}}

\author{Kiyohiro Nakayama$^{1*}$~~~Jan Ackermann$^{1,2*\dagger}$~~~Timur Levent Kesdogan$^{1,2*\dagger}$\\
Yang Zheng$^1$~~~Maria Korosteleva$^2$~~~Olga Sorkine-Hornung$^2$\\
Leonidas J. Guibas$^1$~~~Guandao Yang$^1$~~~Gordon Wetzstein$^1$\\
\\[-0.3cm]
$\phantom{}^1$Stanford University~~~~~$\phantom{}^2$ETH Zürich
\vspace{-1cm}
}

\begin{document}
\input{figures/teaser}
\maketitle
\blfootnote{${}^*$Equal Contribution.}
\blfootnote{${}^\dagger$Work done as a visiting researcher at Stanford.}
\input{sec/0_abstract} 
\input{sec/intro_old}
\input{sec/2_related_works}

\input{sec/3_method}
\input{sec/4_results}

\input{sec/5_conclusion}
\input{sec/6_ack}
{
    \small
    \bibliographystyle{ieeenat_fullname}
    \bibliography{main}
}
\onecolumn

\input{sec/X_suppl}
% \clearpage

\end{document}

%% file: preamble.tex
\usepackage{multirow}
\usepackage{amssymb}% http://ctan.org/pkg/amssymb
\usepackage{pifont}% http://ctan.org/pkg/pifont

\newcommand{\pannelstart}[0]{\texttt{<SoP>}}
\newcommand{\pannelend}[0]{\texttt{<EoP>}}
\newcommand{\garmentstart}[0]{\texttt{<SoG>}}
\newcommand{\garmentend}[0]{\texttt{<EoG>}}

\newcommand{\linetok}[0]{\texttt{<L>}}
\newcommand{\quadtok}[0]{\texttt{<Q>}}
\newcommand{\cubictok}[0]{\texttt{<B>}}
\newcommand{\arctok}[0]{\texttt{<A>}}

\newcommand{\clinetok}[0]{\texttt{<cL>}}
\newcommand{\cquadtok}[0]{\texttt{<cQ>}}
\newcommand{\ccubictok}[0]{\texttt{<cB>}}
\newcommand{\carctok}[0]{\texttt{<cA>}}

%% file: figures/teaser.tex
\twocolumn[{
\renewcommand\twocolumn[1][]{#1}%
\maketitle
% \fbox{\rule{0pt}{2in} \rule{0.975\linewidth}{0pt}}
\begin{center}
    \centering
    \includegraphics[width=\textwidth]{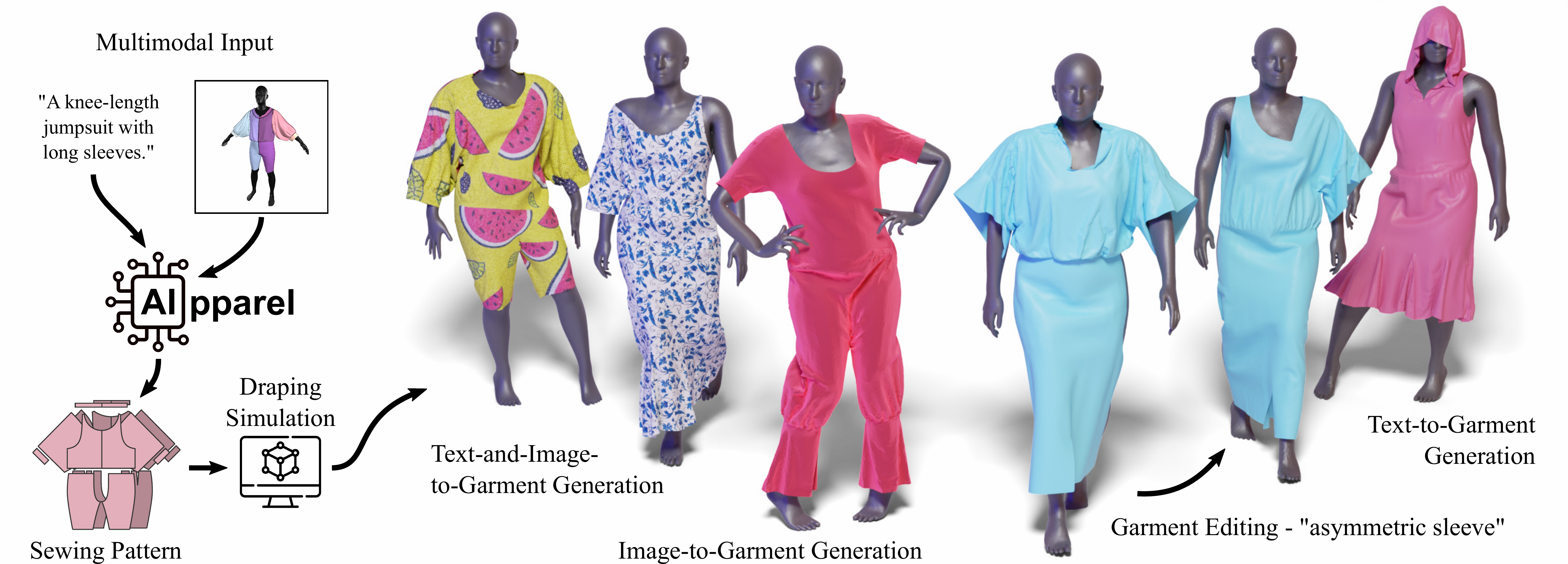}
    \vspace{-1em}
    \captionof{figure}{\textbf{\methodname.} We present a multimodal foundation model for digital garments trained by fine-tuning a large multimodal model on a custom sewing pattern dataset using a novel tokenization scheme for these patterns. \methodname generates complex, diverse, high-quality sewing patterns based on multimodal inputs, such as text and images, and it unlocks new applications such as language-instructed sewing pattern editing. 
    The generated sewing patterns can be directly used to simulate the corresponding 3D garments. 
    %Lastly, our model also enables language-instructed sewing pattern editing. 
    }
    % \srinath{Label `Motion\&Contact CVAE' box above as HuMoR}}
    \label{fig:teaser}
\end{center}%
}]

%% file: sec/0_abstract.tex
% \begin{abstract}
% Garments are an essential aspect of human life. 
% Automatically create fabricable garments, however, remains difficult despite recent progress in generating text, images, and video.
% % This is partially because it involves iteratively predicting and refining sewing patterns based on multimodal inputs like language and imagery. 
% A key challenge of generating fabricable garments is creating and refining sewing patterns based on multimodal inputs such as languages and images.
% To address this challenge, we introduce \methodname, the first multimodal foundation model for sewing patterns that can understand and predict them from diverse inputs. 
% We build a large-scale sewing pattern dataset containing over 120,000 unique samples with multimodal annotations. 
% We propose a novel tokenization scheme to encode complex sewing patterns efficiently. 
% Finally, we train \methodname by fine-tuning state-of-the-art large multimodal models using our dataset and tokenizer. 
% Our method achieves state-of-the-art performance in single-modal sewing pattern prediction tasks.
% In addition, \methodname enables new applications, such as multimodal garment prediction, interactive garment reasoning, and garment editing.
% \end{abstract}
\vspace{-0.3cm}
\begin{abstract}
Apparel is essential to human life, offering protection, mirroring cultural identities, and showcasing personal style. Yet, the creation of garments remains a time-consuming process, largely due to the manual work involved in designing them. To simplify this process, we introduce \methodname, a multimodal foundation model for generating and editing sewing patterns. Our model fine-tunes state-of-the-art large multimodal models (LMMs) on a custom-curated large-scale dataset of over 120,000 unique garments, each with multimodal annotations including text, images, and sewing patterns. Additionally, we propose a novel tokenization scheme that concisely encodes these complex sewing patterns so that LLMs can learn to predict them efficiently. \methodname achieves state-of-the-art performance in single-modal tasks, including text-to-garment and image-to-garment prediction, and enables novel multimodal garment generation applications such as interactive garment editing. The project website is at \url{https://georgenakayama.github.io/AIpparel/}.
\end{abstract}

% Digital sewing pattern generation is a growing research topic and is at the core of many applications, such as virtual try-ons, digital fabrication, and digital avatars. While previous works designed individual models to recover sewing patterns from images, texts, and point-clouds, a typical garment design process often requires an iterative design process using diverse inputs in combination. To address this gap, we introduce \methodname, a multimodal foundation model for digital garment generation. Our work can predict sewing patterns from diverse inputs such as \george{Not sure yet: languages, images, and body shapes} in combination. Furthermore, \methodname also allows for reasoning and editing of sewing patterns required for capturing the complex designer’s intentions. We extend multimodal large language models (LLMs) by expanding their vocabulary with sewing pattern representations. We train on our custom dataset \george{name} -- a large, Q\&A dataset of sewing patterns with multimodal conditioning. We show that \methodname achieves state-of-the-art results in sewing pattern recovery and it enables many novel applications such as interactive garment reasoning and garment editing. Code will be made public upon paper acceptance. 

%% file: sec/intro_old.tex
\section{Introduction}
\label{sec:intro}

%-------------------------------------------------------------------------
Clothing plays a crucial role in society, providing protection from the weather, reflecting societal norms, and serving as a means of personal expression.
% Clothing is essential to human life, offering protection, mirroring cultural identities, and showcasing personal style.
A key stage in garment production is the development of sewing patterns---a set of flat 2D panels with standardized assembly instructions that form a complete 3D garment~\citep{Armstrong:2009}.
%~\cite{liu2023sewformer, clo3d}.
% Pattern making is a time-consuming and challenging task because sewing patterns often involve complex geometry, requiring even experienced tailors to undergo multiple iterations, incorporating feedback from different modalities, including language descriptions of fit and feel of the garment and visual references of its appearance. 
Pattern making is a challenging task due to the complex geometric relationship between the 2D pattern and the draped 3D shape of the sewn garment. 
Even an experienced tailor must go through multiple iterations, incorporating feedback from various sources, including verbal descriptions of the garment's fit and feel, as well as visual references of its appearance. 
% because sewing patterns often involve complex geometry, requiring even experienced tailors to undergo multiple iterations, incorporating feedback from different modalities, including language descriptions of fit and feel of the garment and visual references of its appearance. 
%In this paper, we explore the question: can we develop algorithms to predict and generate sewing patterns from different modalities?
To simplify the pattern-making process, we explore strategies to leverage emerging generative models with multimodal input, such as text and images.
% how one can learn to generate sewing patterns from 
% using large multimodal generative models.

State-of-the-art sewing pattern prediction methods are designed to work with one specific input modality, such as 3D points~\cite{NeuralTailor2022,bang2021estimating,chen2015_garment_modeling_w_depth}, images~\cite{pietroni2022computational,yunchu2007prototype,wolff20213d,nariatposeindependent,yang2016detailedgarmentrecoverysingleview}, or language~\cite{he2024dresscodeautoregressivelysewinggenerating}.
While effective within their respective domains, these single-modal approaches are often challenging to adapt to garment prediction tasks requiring different or combined input modalities. 
Expanding these methods to multimodal pattern prediction presents two primary challenges. 
First, no large-scale multimodal sewing pattern dataset is publicly available. 
Second, the capacity to accurately interpret multimodal inputs typically only emerges in large models with billions of parameters~\cite{achiam2023gpt,liu2023llava}. 
It remains uncertain how to efficiently scale existing methods to models of this size.
% Current multimodal garment prediction efforts generally focus only on generating garment images~\cite{baldrati2023multimodal,jiang2022text2human}. 
% Converting these images into sewing patterns typically involves an additional stage operating independently of the original multimodal input, which can be error-prone.
% Prior methods capable of generating sewing patterns from multimodal inputs, such as \citet{Wang2018LearningAS}, are usually restricted to garments with a predefined set of parameters, limiting their applicability to garments with complex geometries.

% Rather than training specialized models from scratch for pattern prediction from a single modality, we
In this paper, we propose to build such multimodal garment generative models by extending an existing large multimodal model (LMM)~\cite{gpt4tools, touvron2023llama2openfoundation, liu2023llava} to understand sewing patterns with complex geometries.
To achieve this, we annotate the largest sewing pattern dataset \cite{korosteleva2024garmentcodedatadataset3dmadetomeasure} with multimodal labels. 
Our annotated dataset is ten times larger than those used by previous state-of-the-art generative methods~\cite{NeuralTailor2022, liu2023sewformer, he2024dresscodeautoregressivelysewinggenerating}, including over 120,000 unique sewing patterns paired with detailed text descriptions, images, edited sewing patterns, and editing instructions.
Fine-tuning an LMM to perform multimodal garment generation requires representing complex garments in a format that LMMs can understand---tokens.
For this task, we develop a novel tokenization method that is both expressive in representing the complex geometries of sewing patterns and concise enough to fit within the limited context length of existing LMMs, making fine-tuning computationally efficient.

Combining these components, we present \methodname, a large multimodal model for generating sewing patterns.
\methodname can predict sewing patterns with complex geometries and outperforms state-of-the-art methods in single-modal garment prediction, often by a large margin. Moreover, our approach unlocks entirely new multimodal garment generation tasks.
% More importantly, \methodname unlocks multimodal sewing pattern tasks that were previously not possible, \todo{such as language-driven editing of complex garments.}
% and multimodal garment understanding.
Our contributions include:
\begin{itemize}
\item We present GCD-MM, a multimodal sewing pattern dataset extending the largest public dataset of sewing patterns with multimodal annotations.
% We curate the first multimodal sewing pattern dataset of 120,000 unique sewing patterns, each paired with rich multimodal annotations. 
We plan to publicly release the dataset to inspire innovative garment prediction capabilities and further enable research in multimodal garment generation.
\item We develop a novel tokenization scheme and a new training objective for fine-tuning LMMs to predict sewing patterns. 
This tokenization method is critical for retargeting LMMs to multimodal garment prediction tasks efficiently.
% \item We present \methodname, the first large multimodal generative model for sewing pattern prediction capable of taking language, images, and sewing patterns as input.
% We will publish the model and code.

\item We present \methodname, the first multimodal foundation model for sewing pattern prediction capable of taking language, images, and sewing patterns as input.
% To the best of our knowledge, 
% \methodname is the first model capable of generating and editing garments based on multimodal inputs, including natural language and images. 
\end{itemize}

%% file: sec/2_related_works.tex
\section{Related Works}

\paragraph{Garment Generation.}
%%%%%%%%%%%%%%%%%%%%%%%%%%%%%%%%%%%%%%%%%%%%%%%%%%%%%%%%%%
% OUtline:
% 1. Garment generation use different format: images, 3D meshes. 
%   Cons: Many of these format is not predicting sewing pattern, not easy to work with (the argument for needing to output sewing pattern)
% 2. There are many methods predicting sewing patterns from different modalities, such as 3D-scans-to-sewing-pattern~\cite{}, image-to-sewing-pattern~\cite{}, and language-to-sewing-patterns~\cite{}.
% Methods varies from retrieval based~\cite{}, to learning based~\cite{}.
%   Cons: Single modaliy, hard to extend to inputting combination of different modalities. Largely due to 1) need to change the model that can understand complicated signals of different modality format (like language,image,3D, etc); and 2) need a dataset that contains multimodal annotations. 
% 3. There are very few works that take exploratory step in taking multimodal input to predict garments.~\cite{};
%   Cons: fixed template; fixed gamrnet class; cannot take natural language as input; as a result, they can't take complex geometries and take signals such as language as input.
% Our paper: 1) build the dataset; and 2) fine-tune foundation model; and 3) efficient tokenizer for complex geometries - to achieve these three.
%%%%%%%%%%%%%%%%%%%%%%%%%%%%%%%%%%%%%%%%%%%%%%%%%%%%%%%%%%
Prior works have studied learning-based garment generation represented in various formats, including images~\cite{baldrati2023multimodal,jiang2022text2human}, 3D meshes~\cite{srivastava2025wordrobe,de2023drapenet,moon20223d,luo2021garmatnet,jiang2020bcnet,tiwari2021deepdraper,patel2020tailornet,zhu2020deep,ma2020learning,saito2019pifu,zhu2022registering,zheng2024physavatar,li2024garmentdreamer,zheng2024design2cloth}, and sewing patterns~\cite{liu2023sewformer,NeuralTailor2022,he2024dresscodeautoregressivelysewinggenerating,Wang2018LearningAS,pietroni2022computational}.
Our paper focuses on generating sewing patterns, which, compared to other representations, are the industry standard and can be directly used for downstream simulation and manufacturing.
Earlier works have explored a variety of different ways to generate and predict sewing patterns, including retrieval-based methods~\cite{chen2015_garment_modeling_w_depth, hasler_reverse_engineering_garments}, predicting sewing pattern templates with few parameters~\cite{jeong_2015_garment_capture_from_photo, yang_2018_physics_inspired_garment_from_image, WANG2003659,Wang2005DesignAF}, or cutting 3D scans into 2D panels~\cite{daanen_2008,Decaudin_2006,LIU2018113,MENG201268,wang_2009,Sharp:2018:VSC}.
These algorithms usually require heuristics, such as the output garment templates. This limits their flexibility in extending to different input modalities or more complex garment types.
% Leveraging emerging large-scale sewing pattern datasets~\cite{KorostelevaGarmentData,korosteleva2023garmentcode,korosteleva2024garmentcodedatadataset3dmadetomeasure}, 
Further, researchers have successfully applied deep learning methods to generate sewing patterns~\cite{NeuralTailor2022,liu2023sewformer,he2024dresscodeautoregressivelysewinggenerating}.
While these methods can predict accurate sewing patterns based on input conditioning, they are task-specific models designed to work well only in a single modality. 
% For example, NeuralTailor~\cite{NeuralTailor2022} focus on point cloud input, Sewformer~\cite{liu2023sewformer} focuses on image-conditioned garment generation, and DressCode focuses on language-to-garment task.
Extending these single-modal methods to a novel modality is difficult, in part because of the lack of large-scale multimodal sewing-pattern datasets and the requirement to redesign the network architecture.
While \citet{Wang2018LearningAS} can predict sewing patterns from multiple modalities, including images, 3D garments, and body measurements, their method is limited to predicting simple garments with a predefined set of parameters.
In this paper, we aim to tackle the challenge of creating a large multimodal generative model by curating the first multimodal garment dataset with complex garment geometries and providing a scalable recipe building on existing large multimodal models.

\vspace{-1em}\paragraph{Extending Large Multimodal Models.}
Large multimodal models have gained significant attention for their ability to understand language and images~\cite{openai2024gpt4technicalreport,anthropic2024claude,vicuna2023, touvron2023llama2openfoundation, alpaca}.
Efforts to extend LMMs to additional domains typically fall into two categories. 
Optimization-free approaches~\cite{2023interngpt, hugginggpt,wu2023visualchatgpttalkingdrawing, huang2023audiogptunderstandinggeneratingspeech, gpt4tools, wang2023visionllmlargelanguagemodel, yang2023mmreact} employ prompt engineering. The other option is to fine-tune LMMs to take the new modality as input and/or output.
The latter approach was first introduced for vision-language models~\cite{liu2023llava, zhu2023minigpt, tong2024cambrian1} and subsequent works extended it to other modalities~\cite{zhang2023speechgpt, feng2024chatpose,lai2023lisa,xu2023pointllm,rt22023arxiv,2023videochat}. 
Their approaches typically involve using pre-trained encoders~\cite{xu2023pointllm, lai2023lisa, 2023videochat} or standard discrete representations~\cite{zhang2023speechgpt, rt22023arxiv} to convert the input modalities into tokens and align them with the text feature space of the LMMs. In particular, LLaVA~\cite{liu2023llava} is pioneering in fine-tuning Large Language Models (LLMs) for visual understanding. It uses a pre-trained vision encoder to encode images into tokens, and a trainable projection layer to project the visual tokens into the LLM's feature space. We build our work on top of LLaVA by fine-tuning it to understand sewing patterns. This presents unique challenges, however, due to the lack of pretrained encoders or learning-efficient representations for sewing patterns. 
This motivates us to design an efficient, learning-friendly tokenizer and a fine-tuning objective for sewing pattern prediction.

% \begin{figure}
%     \centering
%     \includegraphics[width=\linewidth]{example-image-a}
%     \caption{\textbf{Sampled Data from GCD-MM Dataset}}
%     \label{fig:enter-label}
% \end{figure}
\vspace{-1em}\paragraph{Garment Datasets.}
Garment datasets mostly fall into one of the following three categories: 1) datasets based on 3D scans of real-world garments~\cite{antic2024close,ho2023learning,bhatnagar2019multi,ma2020learning,tiwari2020sizer,xu2023clothpose,zhu2020deep}, 2) datasets of designer-created garments~\cite{black2023bedlam,zou2023cloth4d}, and 3) datasets containing mostly procedurally generated sewing patterns~\cite{bertiche2020cloth3d,jiang2020bcnet,KorostelevaGarmentData,liu2023sewformer,narain2012adaptive,pumarola20193dpeople,vidaurre2020fully,wang2011data,Wang2018LearningAS}.
While 3D garment scans and designer-created garments can accurately capture the real-world complexity of garments, they are expensive to obtain, which limits the scale of these categories of data.
Our work focuses on leveraging large-scale procedurally generated sewing pattern datasets.
% curate a large-scale multimodal garment dataset through synthetic data.
% Existing synthetic datasets can be limited in the diversity of garment styles ~\cite{narain2012adaptive,wang2011data,Wang2018LearningAS}, available labels~\cite{bertiche2020cloth3d,jiang2020bcnet,pumarola20193dpeople,zhou2023clothesnet}, or fitted body-type~\cite{KorostelevaGarmentData}.
To the best of our knowledge, the largest synthetic sewing pattern datasets available are DressCode~\cite{he2024dresscodeautoregressivelysewinggenerating}, SewFactory~\cite{liu2023sewformer}, and GarmentCodeData (GCD)~\cite{korosteleva2023garmentcode,korosteleva2024garmentcodedatadataset3dmadetomeasure}. 
None of their annotations, however, contain the full combinations of text, images, and sewing pattern edits, making them insufficient for training a multimodal sewing pattern generative model. 
% Adding additional modalities like texts is essential for multimodal sewing pattern generation~\cite{he2024dresscodeautoregressivelysewinggenerating,dalle-3}.
To overcome this data gap, we curate the first large-scale multimodal sewing pattern dataset by expanding GCD with annotations including text, editing pairs, and editing instructions. Tab.~\ref{tab:datasets} compares different sewing pattern datasets and their annotation modalities.
\input{tables/datasets}

%% file: tables/datasets.tex
\begin{table}[t]
    \centering
    \begin{tabular}{@{}lcccc@{}}
    \toprule
    Dataset & Total & Text & Image & Edits \\
    \midrule
    \citet{Wang2018LearningAS} & 8k & \xmark & \cmark & \cmark \\
    \citet{KorostelevaGarmentData} & 23.5k & \xmark & \cmark & \xmark\\ 
    Sewfactory~\cite{liu2023sewformer} & 19.1k & \xmark & \cmark & \xmark \\
    DressCode~\cite{he2024dresscodeautoregressivelysewinggenerating} & 20.3k & \cmark & \xmark & \xmark \\
    GCD~\cite{korosteleva2024garmentcodedatadataset3dmadetomeasure} & 130k & \xmark & \cmark &  \xmark \\
    \textbf{GCD-MM (Ours)} & 120k & \cmark & \cmark & \cmark \\
    \bottomrule
    \end{tabular}
    \vspace{-0.75em}
    \caption{
    \textbf{Modalities of Sewing Patter Datasets. }
    GCD-MM is a large-scale sewing pattern dataset with multimodal annotations, including text, images, and edited patterns.
    % Our dataset annotates GarmentCodeData~\cite{korosteleva2024garmentcodedatadataset3dmadetomeasure} with multimodal annotations including both text, images and edited patterns.
    }
    \label{tab:datasets}
    \vspace{-1em}
\end{table}

%% file: sec/3_method.tex
\section{Method}\label{sec:method}

We propose a large multimodal generative model for sewing patterns by fine-tuning existing LMMs on a multimodal sewing pattern dataset. For this purpose, we first curate a sewing pattern dataset with multimodal annotations (\cref{sec:data-curation}). We then describe how to train our model, \textit{\methodname}, using an efficient tokenization scheme for sewing patterns, with LlaVA 1.5-7B~\cite{liu2023llava} as a base model (\cref{sec:training}). 

\subsection{Multimodal GarmentCode Dataset}\label{sec:data-curation}

We create annotations covering many modalities to train a multimodal sewing pattern generative model. Specifically, we build on top of the largest existing sewing-pattern dataset, GarmentCodeData (GCD)~\cite{korosteleva2024garmentcodedatadataset3dmadetomeasure}, to incorporate two other modalities: text descriptions and sewing pattern pairs with editing instructions. We dub our dataset GarmentCodeData-MultiModal (GCD-MM).

\vspace{-1em}\paragraph{Text description of sewing patterns.}
To enable applications such as text-conditioned sewing pattern generation, it is important to obtain detailed text annotation describing the sewing patterns~\cite{he2024dresscodeautoregressivelysewinggenerating,dalle-3}.
\citet{he2024dresscodeautoregressivelysewinggenerating} created short keyword descriptions of sewing patterns by prompting GPT4V with rendered images. However, this method suffers from hallucination, and the short keywords are insufficient to describe the garments in detail, leading to ambiguities.
Our pipeline improves on this by leveraging the design parameters associated with each synthetically generated sewing pattern to create accurate descriptions that capture the garment's key features.
Specifically, we develop a rule-based algorithm to generate a set of short phrases, including a garment type (e.g., ``midi dress'', ``godet skirt'') and brief descriptions based on distinctive characteristics (e.g., ``flared hem'', ``V-neckline''). 
To obtain the final sewing pattern description, we prompt GPT-4o~\cite{gpt4tools} using the rule-based short captions and the rendered views of the draped garment. 
Our approach reduces GPT-4o's hallucination and results in more accurate descriptions in natural language.
Please refer to the supplementary for caption comparison with DressCode and the prompts and rules we used to generate them.
% \jan{Another key part is that this can generate characteristics that are not even part of the rule-based approach}
% \todo{Some statistics: how many queries do we use? how much does it cost? refer to the supplementary for some evidences that our prompt is better?}
% This approach enhances the descriptions with conversational, sentence-like captions, combining both rule-based and generative techniques to prevent hallucinations while capturing more nuanced characteristics. 
% For editing instructions, we generate a standardized set of 56 descriptive edits based on our rule-based framework.

% \noindent\textbf{Language-instructed Sewing Pattern Editing.}
\vspace{-1em}\paragraph{Language-instructed Sewing Pattern Editing.}
We also augment GCD with language-instructed editing annotations. Specifically, we use the programming abstraction from GarmentCode~\cite{korosteleva2023garmentcode} to create paired sewing patterns with corresponding text instructions describing the applied edits. 
We first manually specify a series of sewing pattern edits using the abstraction. This includes edits such as adjustments in skirt and pants length, changing insert and neckline styles, and adding or excluding a hood or sleeve. 
For each modification, we generate captions using a text template to describe the applied changes.
See the supplementary for editing templates and captions examples.
% \todo{Need statistics; how many of what kind of edits.}

\input{figures/method}

\subsection{\methodname}\label{sec:training}
\methodname fine-tunes LLaVA 1.5-7B on our GCD-MM dataset to generate sewing patterns from multimodal input. 
For this purpose, we need to encode sewing patterns into a compact list of tokens for LLaVA's input.
% To build efficient sewing pattern representation as tokens, we borrow from SVG generative works~\cite{carlier2020deepsvg} to represent sewing patterns as drawing instructions. 
% To preserve full precision of the continuous parameters, 
We also propose a novel fine-tuning objective that allows \methodname to generate both discrete tokens and continuous parameters. 
\Cref{fig:pattern_tok} shows an overview of our method. 
% \subsection{Sewing Pattern Tokenization}\label{sec:tokenizer}
% To fine-tune LMMs on our GCD-MM dataset, we represent sewing patterns as tokens that transformers can generate. In the following sections, we first precisely define sewing patterns as a set of 2D flat surfaces in 3D with stitching information. Then, we describe our novel sewing pattern tokenization scheme that efficiently encodes sewing patterns for transformers without loss of precision.  
% \label{sec:pattern_tok}

\vspace{-1em}\paragraph{Pattern Representation.}
% \subsubsection{Sewing Pattern Representation}
Following GCD~\cite{korosteleva2024garmentcodedatadataset3dmadetomeasure}, we define sewing patterns as a set of 2D panels in 3D with stitching information. 
A sewing pattern $\mathcal{P} = (P, S)$ is a tuple consisting of $N$ panels $P = \set{P_1, \dots, P_N}$ and stitching information $S$. 
Each panel $P_i$ is a planar surface with vertices $V_i = \{v^{(i)}_1, \dots, v^{(i)}_{n_i}\}$ and edges $E_i = (e^{(i)}_1, \dots, e^{(i)}_{n_i})$, where each edge contains two endpoints connecting $(v_k^{(i)}, v_{k'}^{(i)})$ with $k' = k \mod n_i + 1$.
Since each panel is defined in its own coordinate frame, we always set $v^{(i)}_1 = 0\in \R^2$. 
An edge can be a straight line, a quadratic or cubic Bézier curve, or an arc, and includes its corresponding control vertices $\bm{c}^{(i)}_k$.
% (zero for lines, one for quadratic Bézier curves and arcs, and two for cubic Bézier curve). 
Each panel also includes a rigid 3D transformation $R$ that transforms $P_i$ into the global coordinate frame for draping. 
Lastly, each panel contains a unique name indicating the panel type for the designers.
We define stitching information $S$ as a set of edge pairs among panel edges, i.e., $S = \{(e^{(i_1)}_{k_1}, e^{(j_1)}_{l_1}), \dots, (e^{(i_m)}_{k_m}, e^{(j_m)}_{l_m})\}$ where each $(e^{(i_s)}_{k_s}, e^{(j_s)}_{l_s})$ indicates that edge $e^{(i_s)}_{k_s}$ from panel $P_{i_s}$ will be stitched with edge $e^{(j_s)}_{l_s}$. See the supplementary for representation details.

\vspace{-1em}\paragraph{Sewing Pattern Tokenization.}
% \subsubsection{Sewing Pattern Tokenization}
The sewing pattern representation in GCD contains both continuous parameters, such as panel vertex coordinates, and discrete parameters, such as the number of panels and stitches.
% including the panel vertices, control points, and transformations, as well as discrete ones, such as the number of panels, edges, and stitches. 
This poses challenges in representing each sewing pattern compactly as a set of tokens for the transformer's prediction. 
Prior works rely on extensive zero-padding to ensure that all sewing patterns can be represented as a fixed-length vector~\cite{liu2023sewformer, NeuralTailor2022, he2024dresscodeautoregressivelysewinggenerating}.
This approach is impractical for the complex sewing patterns found in GCD-MM, as it produces an excessively long context. For example, the tokenization scheme of \citet{he2024dresscodeautoregressivelysewinggenerating} requires more than 30k tokens to represent a typical sewing pattern in the GCD-MM dataset, making it extremely inefficient for generation and learning.\footnote{See supplementary for a detailed analysis.}
% One solution is to use zero padding to make sure that all sewing patterns can be represented as a long vector of the same length. While this is the approach adopted by previous works~\cite{liu2023sewformer, NeuralTailor2022, he2024dresscodeautoregressivelysewinggenerating} on previous datasets with simple patterns, it becomes impractical for learning meaningful representation for complex sewing patterns present in GCD-MM. 

% To handle the diverse sewing patterns in GCD-MM, 
Inspired by recent work on vector graphics generation~\cite{carlier2020deepsvg}, we develop a tokenization scheme that efficiently represents sewing patterns as a sequence of drawing commands. 
% Instead of zero-padding, 
Specifically, we introduce four special tokens to indicate the start of a garment (\garmentstart) and the end of a garment (\garmentend), as well as the start of a panel (\pannelstart) and the end of a panel (\pannelend).
% Using these tokens, $\texttt{<PAN S>}\cdots\texttt{<PAN E>}$ represents the context of each panel and $\texttt{<PAT S>}\cdots\texttt{<PAt E>}$ represents the context of the sewing pattern. 
With these tokens, each sewing pattern can be represented as 
% \begin{align}
\begin{equation}
\label{eq:pat_tok}
% \begin{split}
    \operatorname{E}_g(\mathcal{P}) =
    \garmentstart\text{E}_p(P_1, S)\cdots\text{E}_p(P_n, S)\garmentend,
    % \\
    % \text{E}_p(P, S) &= \texttt{\pannelstart}\text{E}(P_i, S)\texttt{\pannelend} \\
% \end{split},
\end{equation}
where $\operatorname{E}_p$ tokenizes panel $P$ in the form of $\pannelstart \dots \pannelend$.
% This sequence contains three information:
% \end{align}
$\operatorname{E}_p$ consists of three pieces of panel information: name, transformation, and edges. 
The panel name is tokenized using LLaVA-1.5-7B's text tokenizer and inserted after \pannelstart. 
We introduce a new token \texttt{<R>} and place it after the panel name to represent the panel's transformation. 
Each edge type also corresponds to two special tokens, depending on whether the edge ends at the starting endpoint: line (\linetok, \clinetok), quadratic Bézier curve (\quadtok, \cquadtok), cubic Bézier curve (\cubictok, \ccubictok), and arc (\arctok, \carctok). 
% Because the last edge always ends with $0\in \R^2$, we distinguish the last edge from the rest by assigning it with a token indicating the closure of the edge type: closure of line (\texttt{<C_L>}), quadratic Bézier curve (\texttt{<C_QC>}), cubic  Bézier curve (\texttt{<C_CC>}), and arc (\texttt{<C_A>}). 
% Therefore, in total, there are eight types of edge-type tokens. 
We also introduce a set of stitching tag tokens $\set{\texttt{<t1>}, \dots, \texttt{<tM>}, \texttt{<tN>}}$ to represent stitching information $S$. 
We associate each edge with a stitching tag so that $(e^{(i_s)}_{k_s}, e^{(j_s)}_{l_s}) \in S$ iff there exists $a \in \set{1, \dots, M}$ such that $e^{(i_s)}_{k_s}$ and $e^{(j_s)}_{l_s}$ are both associated with \texttt{<T$a$>}. 
If an edge is not stitched to another edge, it is associated with the null tag $\texttt{<tN>}$. 
% We set $M=108$ to be the maximum number of stitches in GCD-MM. 
For example, a panel consisting of two lines stitched together, one cubic Bézier curve and an arc is tokenized as 
\begin{equation*}
% \label{eq:pan_tok}
    \begin{split}
        % &\texttt{<PAN S>}\text{tokenize}(P_i)\texttt{<PAN E>} =\\
    &\pannelstart\text{[panel name]}\texttt{<R>}\linetok\texttt{<t1>}\linetok\texttt{<t1>}\\
    &\cubictok\texttt{<tN>}\carctok\texttt{<tN>}\pannelend.
    \end{split}
\end{equation*}
% Here, the first two lines are stitched together because they share the same stitching tag. 
Compared to the DressCode tokenizer~\cite{he2024dresscodeautoregressivelysewinggenerating}, our proposed scheme uses around 100 times fewer tokens to describe the same garment. On average, we represent a sewing pattern with around 250 tokens with a maximum of 838 tokens on GCD-MM, whereas DressCode uses more than 30k tokens for each sewing pattern on the same data. 

\input{figures/image2garment}
\input{tables/image2garment}

\vspace{-1em}\paragraph{Notation.} From now on, we use bold letters (e.g., $\bm{X}\in \R^{N \times D}$) to denote the input embedding sequence to the transformer. We denote the $i$-th embedding in $\bm{X}$ as $\bm{X}_i$, and $\bm{X}_{<i}, \bm{X}_{>i}$ are sliced sequences before or after the $i$-the embedding, respectively. We use $f_\phi$ to denote the language transformer from LLaVA. We use $\bm{X}$ to denote tokens before passing through $f_\phi$ and $\bm{H} = f_\phi(\bm{X})$ as the output hidden features from the transformer. 

\vspace{-1em}\paragraph{Continuous Parameters.} 
The tokenization scheme in Eq.~\ref{eq:pat_tok} does not include any continuous parameters such as vertex positions, control points for edges, or rigid transformation of panels. 
Prior works represent continuous parameters as quantized tokens in discrete space~\cite{he2024dresscodeautoregressivelysewinggenerating,
nash2020polygenautoregressivegenerativemodel}.
This introduces quantization error for the continuous parameters and uses more tokens per panel, leading to a longer, inefficient representation. 
Inspired by recent approaches of extending LMMs~\cite{feng2024chatpose, lai2023lisa}, we propose using small regression heads to map hidden features of the transformer to the continuous parameters. 
Specifically, we define an MLP $g^{(\text{e})}_\theta: \R^D \to \R^C$ to map LLaVA's hidden features from the last layer to vertices and control points. 
% We set $C = 8$ and use the first two channels as vertex regression, the next four as the control points to the quadratic and cubic Bézier curves, and the last two channels as the additional vertex for representing an arc. 
As illustrated in Fig.~\ref{fig:pattern_tok}, $g^{(\text{e})}_\theta$ takes the output embedding corresponding to the token right before the edge type token. Concretely, if the $i$-th token, $\bm{X}_i$, corresponds to an edge-type token for edge $e$, its associated output embedding $\bm{H}_{<i} = f_\phi(\bm{X}_{<i}) \in \R^{(i-1)\times D}$ is used to predict $e$'s endpoint and control parameters via $g^{(\text{e})}_\theta(\bm{H}_{i-1})$. Similarly, we also define a transformation regression head $g^{(R)}_{\theta^\prime}: \R^D \to \R^7$ mapping the hidden features of \texttt{<R>}'s previous token to a translation $T \in \R^3$ and a rotation quaternion $q \in \mathbb{H}$. In this way, continuous parameters are regressed using small regression heads that are jointly trained with the transformer, leading to more efficient context length usage in representing sewing patterns. At training time, we use ground-truth parameters for supervision, and during generation, we use the last hidden feature from the output for regression should an edge-type token or a transformation token be sampled. 

%\vspace{-1em}
\paragraph{Positional Embeddings.} To include information on continuous parameters in the sewing pattern tokenization defined in Eq.~\ref{eq:pat_tok}, we include the second endpoint of each edge as a positional embedding added to the token embedding. Specifically, we define $h^{(e)}_\varphi: \R^2 \to \R^D$ as a two-layer perceptron. For each edge $e = (v_1, v_2)$ with an edge-type token embedding of $\bm{X}_i$, we add $h^{(e)}_\varphi(v_2)$ to $\bm{X}_i$ to inform the language model $f_\phi$ of the vertex positions. We also define $h^{(R)}_{\varphi^\prime}: \R^7 \to \R^D$ to be the projection function for transformations. The transformation parameters $(t, q) \in \R^7$ for each panel are projected using $h^{(R)}_\varphi$ and added to the token embedding for \texttt{<R>}. At training time, we use ground-truth vertex and 3D transformations as positional embeddings, and during generation, we use the parameters predicted by the regression heads. 

\paragraph{Training.} We keep both the vision encoder and projection frozen, and fine-tune all weights in the language model $f_\phi,$ regression heads $g^{(\text{e})}_\theta, g^{(R)}_{\theta^\prime}$, and the positional embedding projection layers $h^{(\text{e})}_\varphi, h^{(R)}_{\varphi^\prime}$. The fine-tuning loss is defined as a combination of cross-entropy (CE) on the discrete tokens and $L_2$ loss on the continuous parameters: 
\begin{equation}
    \label{eq:obj}
    \begin{split}
        \mathcal{L} &= \text{CE}(f_\phi(\bm{X}_{<-1}), \bm{X}_{1>}) \\
        &+ \lambda \sum_{e^\prime} \norm{g^{(\text{e})}_\theta\circ f_\phi\paren{\bm{X}_{<i_{e^\prime}}} - v^{e^\prime}_2}_2 \\
        &+ \lambda \sum_{R^\prime}\norm{g^{(R)}_{\theta^\prime}\circ f_\phi\paren{\bm{X}_{<i_{R^\prime}}} - R^\prime}_2.
    \end{split}
\end{equation}
Here, $\sum_{e^\prime}$ is the sum over all edges in the sewing pattern's sequence $\bm{X}$. For each edge $e^\prime$, its second endpoint is denoted as $v^{e^\prime}_2$. Similarly, $\sum_{R^\prime}$ is the sum over all the transformations. We do not explicitly include the positional embedding in Eq.~\ref{eq:obj}, but it is added according to the rules defined in the previous paragraph.

%% file: figures/method.tex
\begin{figure*}[th]
    \centering
    \includegraphics[width=\textwidth]{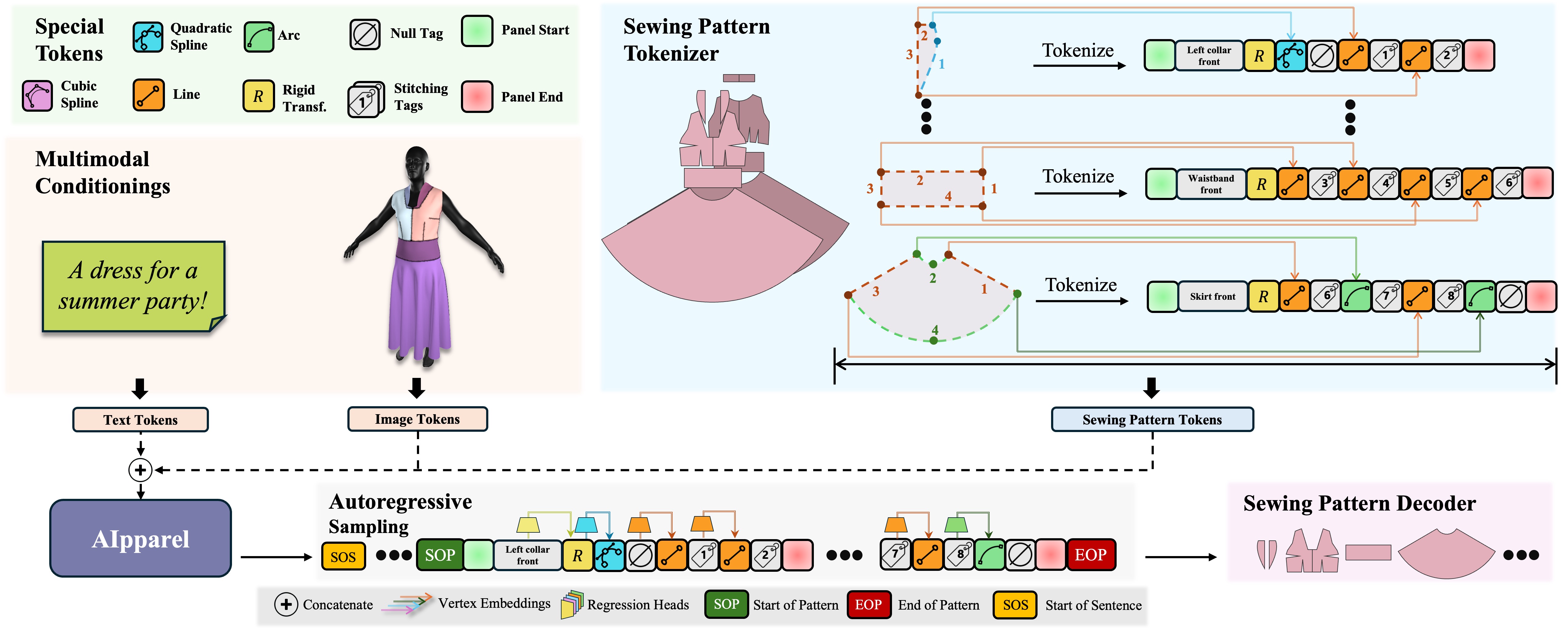}
    \vspace{-1.5em}
    \caption{\textbf{Illustration of Our Method}. AIpparel uses a novel sewing pattern tokenizer (light blue region) to tokenize each panel into a set of special tokens (light green region). Panel vertex positions and 3D transformations are incorporated using positional embeddings (colored arrows) to the tokens. \methodname takes in multimodal inputs, such as images and texts (light orange region), to output sewing patterns using autoregressive sampling (light grey region). Finally, the output is decoded to produce simulation-ready sewing patterns (light pink region). 
    See~\Cref{sec:method} for method details.
    }
    \label{fig:pattern_tok}
\end{figure*}

%% file: figures/image2garment.tex
\begin{figure*}[ht]
    \includegraphics[width=\textwidth]{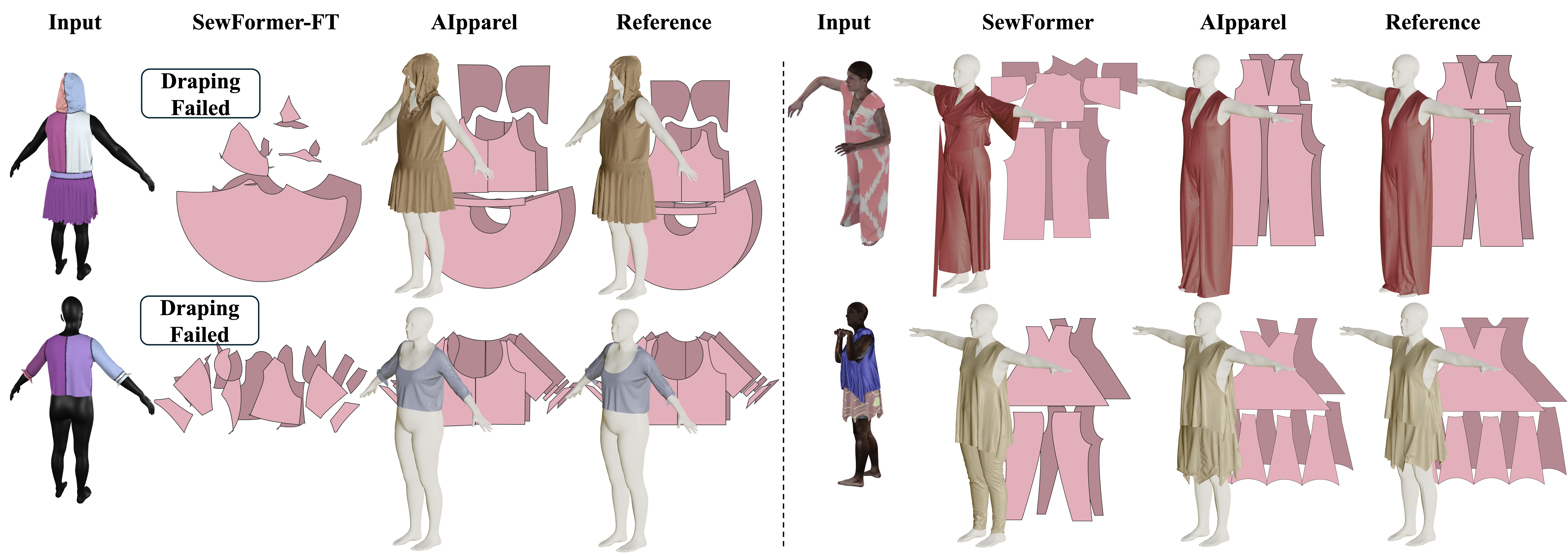}
    \vspace{-0.5em}
    \caption{\textbf{Image-to-Garment Prediction (Qualitative).}
     \textit{GCD-MM (Left)}: our model can reconstruct suitable sewing patterns from the input image alone. In contrast, SewFormer does not produce simulation-ready sewing patterns despite fine-tuning. \textit{SewFactory (Right)}: SewFormer produces inaccurate panels (top row) and incorrect garment types (bottom row) while \methodname accurately recovers sewing patterns from the images, resulting in superior simulation results. See Sec.~\ref{sec:img2garment}.
    }
    \label{fig:image2garment}
\end{figure*}

%% file: tables/image2garment.tex
\begin{table*}[ht]
    \centering
    \begin{tabular}{@{}lc@{}cccccc@{}}
    \toprule
         Dataset & Method &  Panel L2 $(\downarrow)$ & \#Panel Acc $(\uparrow)$ & \#Edge Acc $(\uparrow)$  & Rot L2 $(\downarrow)$ & Transl L2 $(\downarrow)$ & \#Stitch Acc $(\uparrow)$\\\midrule
    % SewFormer pretrained & & & & && \\    
    % DressCode pretrained (+ GPT4V)  & & & & &&\\
    % Sewformer-pretrained \\
    \multirow{2}{*}{Sewfactory}
    & SewFormer & 3.3 & 89.8 & 99.3 & .008 & 0.8 & 99.2\\
    & \textbf{\methodname} & \textbf{2.8} & \textbf{93.9} & \textbf{99.9} & \textbf{.005} & \textbf{0.6} &  \textbf{99.8}\\
    \midrule
    \multirow{2}{*}{GCD-MM}
    & SewFormer-FT & 12.3 & 79.4 & 44.7 & .040 & 4.5 & 2.8\\
    & \textbf{\methodname}   & \textbf{5.4} & \textbf{85.2} & \textbf{82.7} & \textbf{.020} & \textbf{2.7} & \textbf{77.2} \\
    \bottomrule
    \end{tabular}
    \vspace{-0.5em}
    \caption{
    \textbf{Image-to-Garment Prediction (Quantitative).} 
    % We report the reconstruction performance on the Sewfactory and the GarmentCode datasets.
    % For the SewFactory dataset~\cite{liu2023sewformer}, we compare \methodname to the pre-trained SewFormer model.
    % For the GarmentCode dataset~\cite{korosteleva2024garmentcodedatadataset3dmadetomeasure}, we compare \methodname to SewFormer finetuned on the GarmentCode dataset. 
    % We compare to SewFormer trained or finetuned at the same dataset.
    \methodname achieves state-of-the-art performance in both datasets and surpasses SewFormer-FT by a large margin on GCD-MM. 
    % We report the garment reconstruction metrics of both Sewformer and our model in \todo{GarmentCode~\cite{korosteleva2023garmentcode} dataset, which contains complex and real-world garments}. Our model achieves better performance in all metrics.\todo{On Sewformer's dataset itself?}
    }
    \label{tab:image2garment}
\end{table*}

%% file: sec/4_results.tex
\section{Experiments}
\label{sec:experiments}
% \begin{figure}
%     \centering
%     \includegraphics[width=\linewidth]{example-image-a}
%     \caption{\textbf{Pattern Reasoning Dataset}. The dataset consists of tuples of textual descriptions, images, and patterns. The goal is to recover the sewing pattern from the image based on the text description.}
%     \label{fig:complex_data}
% \end{figure}
We validate the effectiveness of our model on multiple tasks and conduct an ablation study on the key technical designs.
\vspace{-1em}\paragraph{Training Details.}
We train \methodname on GCD-MM multimodal data samples for image-to-garment and text-to-garment generation, as well as text-based garment editing. We randomly split GCD-MM into train-validation-test subsets with a 90:5:5 ratio. All of our results on GCD-MM below are predicted using a single model. See the supplementary for a complete implementation and training setup. 

% \noindent\textbf{Metrics.} 
\vspace{-1em}\paragraph{Metrics.} 
To quantitatively measure our sewing pattern predictions, we use reconstruction metrics established by previous works~\cite{liu2023sewformer, NeuralTailor2022}. 
Given a pair of generated and ground-truth sewing patterns, we measure 1) \textit{Panel L2, Rot L2, and Transl L2}: average vertex, rotation and translation L2 distance between predicted and ground-truth panels; 2) \textit{\#Panel Accuracy}: percentage of sewing patterns with correctly predicted number of panels; 3) \textit{\#Edge Accuracy}: percentage of correctly predicted edges in each correctly predicted panel; 4) \textit{\#Stitch Accuracy}: accuracy of predicted stitches compared to ground truth. 
To save space in compact tables, we report \textit{Accuracy}, the product of \textit{\#Panel Accuracy} and \textit{\#Edge Accuracy}, to provide a comprehensive measurement of garment reconstruction quality. 
All L2-based metrics are measured in centimeters except for rotation.
% These metrics provide a comprehensive set of evaluations of the predicted garment quality. 

% \input{figures/SewFormer_comp/fig}
% \timur{Todo, place fig somewhere else if needed}

%%%%%%%%%%%%%%%%%%%%%%%%%%%%%%%%%%%%%%%%%%%%%%%%%%%%%%%%
% Singlemodal garment prediction (Img2Garment)
%%%%%%%%%%%%%%%%%%%%%%%%%%%%%%%%%%%%%%%%%%%%%%%%%%%%%%%%
\subsection{Image to Sewing Pattern Prediction}
\label{sec:img2garment}
We test our model's capability to reconstruct garments from a single image using two datasets: SewFactory~\cite{liu2023sewformer} and GCD-MM. For the baseline, we compare with SewFormer's pre-trained model on the SewFactory dataset. Because SewFormer did not release their train--test split for their pre-trained model, we use a custom split for a fair comparison. For GCD-MM, we fine-tune SewFormer until its validation loss no longer improves. We denote it as \textit{SewFormer-FT}. 

Tab.~\ref{tab:image2garment} shows quantitative comparisons on the two datasets. \methodname outperforms the baselines on both datasets, suggesting that our method outputs more accurate sewing patterns than the baseline. In particular, \methodname shows a large performance improvement over SewFormer-FT on the difficult GCD-MM dataset, indicating the effectiveness of our method in predicting more complex sewing patterns. Fig.~\ref{fig:image2garment} shows qualitative results. The two examples on the left show that SewFormer-FT fails to reconstruct simulatable garments despite fine-tuning. This suggests that SewFormer cannot adapt to complex garments with small panels and diverse edge types. In contrast, our model predicts sewing patterns matching the input images, including small panels such as the waistband on the top row and the sleeve cuffs at the bottom. The two examples on the right show results on SewFactory~\cite{liu2023sewformer}. The pre-trained SewFormer fails to predict the garment in the top row with sleeves and the bottom row's skirt as a pair of pants, while \methodname correctly predicts the sewing patterns based on the inputs. 

\subsection{Multimodal Garment Generation}\label{sec:multi2garment}

We evaluate the effectiveness of \methodname in various multimodal garment generation scenarios. 
Specifically, we assess its performance on a set of 100 garments with 5 types of multimodal inputs (20 test samples each): 1) texts, 2) images, 3) a combination of text and images, 4) open-ended prompts that require reasoning, and 5) editing instructions.
% This benchmark highlight a wide suite of multimodal garment capabilities.
Success in such a benchmark requires the model to make accurate predictions conditioned on a variety of different multimodal input formats, as well as having an understanding of common-sense knowledge. 
Refer to Fig.~\ref{fig:multimodal} for generation examples in these tasks. We abbreviate the text input for compactness. Refer to the supplementary for complete examples. 

Because no existing work handles multimodal sewing pattern generation, we adopt state-of-the-art (SOTA) single-modal generative methods, i.e., SewFormer-FT and DressCode, to perform multimodal tasks. For this purpose, we augment these baselines using multimodal models, e.g., GPT-4o~\cite{gpt4tools} and DALL-E~\cite{dalle-3}, to translate the multimodal inputs to their input domains (i.e., images and short keyword description). To ensure translation accuracy, we manually inspect the results before querying SewFormer-FT and DressCode. We denote them as \textit{Sewformer-FT}$^\dagger$ and \textit{DressCode}$^\dagger$, respectively. In comparison, our model can directly perform all five categories of multimodal tasks without relying on external modules.
% Specifically, we use GPT-4V to translate garment renders into keywords for DressCode.
% We use DALL-E to generate garment renders from texts before inputting them into SewFormer-FT.

\input{figures/multimodal_reasoning}
\input{tables/reasoning_based_pattern_recov}

We report quantitative comparisons in Tab.~\ref{tab:reason_recov}, measured between the reference and predicted sewing patterns. Compared with single-modal baselines, \methodname performs significantly better. Fig.~\ref{fig:multimodal} shows qualitative comparisons. The first row validates the method's reasoning ability by asking for a suitable sewing pattern for a specific occasion (e.g., a ``semi-formal garden party''). We use DressCode$^\dagger$ as our baseline. Notice that DressCode$^\dagger$ generates a mini-skirt that does not match well the description of a ``semi-formal garden party''. Our model outputs a godet skirt that is more appropriate for this occasion. 
The second row shows an example of sewing pattern generation given a combination of image and text. 
SewFormer-FT$^\dagger$ fails to generate a plausible sewing pattern due to the garment's complexity, whereas \methodname reconstructs the complex garment closely following both visual and textual cues, such as the sleeve and skirt length in the image, and the waistline and neckline descriptions in the text.  

\subsection{Language-driven Sewing Pattern Editing}\label{sec:garmentedit}

%%%%%%%%%%%%%%%%%%%%%%%%%%%%%%%%%%%%%%%%%%%%%%%%%%%%%%%%
% guandao's prompt
% Describe the goal of the experiment:
%       - multimodal enables new capability. Editing is one of the example.
% Task description: 
%       - Input:    input is a sewing pattern, and an editing instruction
%       - Output:   an edited sewing pattern that adhere to the input sewing panels and instructions.
% Experiment description:
%       - Dataset: we have created language-driven editing data for our dataset (~ how many); we do train/validation split in XXX ways.
%       - Baseline: we compare ours with a single-modality baseline, using state-of-the-art multimodal models like GPT-4V.
%       - Ours: just the full model.
% Results:
%       - We can achieve better reconstruction accuracy
%       - (qualitative) We better preserve the input sewing pattern's geometries.
%%%%%%%%%%%%%%%%%%%%%%%%%%%%%%%%%%%%%%%%%%%%%%%%%%%%%%%%
We validate \methodname's ability to perform sewing pattern editing. Given a sewing pattern and text-based editing instructions, the model is tasked with editing the pattern according to the prompt without altering the overall style of the garment.
%Specifically, given inputs consists of a sewing pattern and a textual editing instruction, the model is tasked to edit the sewing pattern to satisfy the user intents described in the text without affecting the overall style of the sewing pattern. 
% For this task, we curate a training dataset with around 110k sewing pattern editing examples and use a held-out test-set to evaluate different models' performances in this task.
Since the existing sewing pattern generation models, DressCode and SewFormer, are not designed for this task, we adapt them for editing using a pre-trained InstructPix2Pix~\cite{brooks2022instructpix2pix} and GPT4o~\cite{gpt4tools}\footnote{See supplementary for details}. We denote them as \textit{DressCode*} and \textit{SewFormer*}, respectively. 
% For DressCode, we render the draped sewing pattern before editing and use GPT4o to output short phrases that describe the edited garment with image and editing instruction inputs. 
% For SewFormer, we use pre-trained InstructPix2Pix~\cite{brooks2022instructpix2pix} to modify the draped sewing pattern rendering based on the editing instruction. 
% The image output by InstructPix2Pix is used as the input to SewFormer. 
\input{figures/editing}
\input{tables/editing}

We report quantitative comparisons in Tab.~\ref{tab:editing}.
Our model outperforms the baselines in both metrics by a large margin, indicating that \methodname performs more accurate edits than the baselines while minimally affecting the rest of the sewing patterns. 
Qualitative results are shown in Fig.~\ref{fig:editing}. DressCode produces results visibly deviating from the input garment. For example, DressCode changes the tank top to a full-length dress in the top row and the tight skirt to a flared one in the bottom row. These mistakes arise because DressCode requires external modules to translate the sewing pattern to short keywords for input, losing important information about the original garment style during the process.
% Qualitatively, both baselines find it difficult to reuse the input garments' sewing patterns, partially because it is difficult to describe the original garments in full detail with either language or images. \jan{This needs reformulation}
In contrast, \methodname directly accepts the sewing pattern and textual instructions as input, allowing it to accurately perform the minimal edits required to modify the garment according to the instructions, as demonstrated in both examples.

\subsection{Ablation Study}\label{sec:ablation}
\input{tables/ablation}
We validate our key technical contributions in an ablation study. Specifically, we compare our tokenizer described in Sec.~\ref{sec:method} with the existing tokenization scheme from DressCode~\cite{chen2024dressautomaticphysicallybased}. We also perform an ablation study on our proposed mixed fine-tuning objective in Eq.~\ref{eq:obj}, comparing it with a cross-entropy-only objective (``\textit{Ours w.o.\ reg}''). We use text-to-garment prediction on DressCode's dataset as our ablation task to compare with DressCode's pre-trained model. 
% In this task, models are tasked to accurately recover the sewing pattern given this input caption.
For a fair comparison, we use the same backbone as DressCode and only change the tokenizer and training objectives.
To implement ``\textit{Ours w.o.\ reg.}'', we quantize vertex positions, edge control parameters, and 3D transformations into 256 bins, which are then predicted using next token prediction. See the supplementary for implementation details. 

% For this model, we quantize the continuous parameter space into uniform intervals. 
% where given the first panel, the model needs to complete the rest of the sewing pattern. 
% Furthermore, we also evaluate the effectiveness of our hybrid fine-tuning objective by comparing our tokenizer with the version where all continuous parameters are predicted discretely on the same task.
% \todo{Trianing detail}
% we trained our model using hyperparameter suggested by the nanoGPT repostiroy~\todo{cite}, which conerges at about $12750$ iterations.
% , using learning rate of $6\times 10^{-5}$, batch size of 320. We use a cosine learning rate scheduler with 100 steps of warm up. 
% Our model for a total of 12750 steps.

Tab.~\ref{tab:ablation} shows the ablation results compared to our full model in text-to-garment tasks. The results are averaged over $100$ samples from DressCode's test set.
Compared to DressCode, our tokenizer, both with and without regression loss, significantly improves the reconstruction fidelity, as demonstrated by the large improvement in the metric values.
Furthermore, by using a mixed training objective in Eq.~\ref{eq:obj}, the reconstruction quality of sewing patterns improves significantly, demonstrating the effectiveness of our objective.
In addition to quality improvements, our tokenization drastically accelerates generation (25$\times$ speedup) compared to DressCode, as shown in the same table. The reported times show the average wall-clock time required to generate and decode a single garment in seconds, measured for each method on a single A6000 GPU.
% Our proposed tokenizer also speeds up the decoding time by about $25\times$, thanks to our tokenization scheme's ability to compactly represent complex geometries in tokens.
Fig.~\ref{fig:ablation} displays reconstructed sewing patterns from all three methods. Notice that DressCode's prediction does not accurately reflect the language description (i.e., the top row's skirt is not flared) and shows geometry artifacts (bottom row, boxed region). Meanwhile, our proposed tokenizer and training objective predict garments with the best visual quality and alignment with the textual description.
% These results suggest that our proposed tokenizer and training objectives leads to improvement in performance.

\input{figures/ablation}

%% file: figures/multimodal_reasoning.tex
\begin{figure}[t]
    \includegraphics[width=\linewidth]{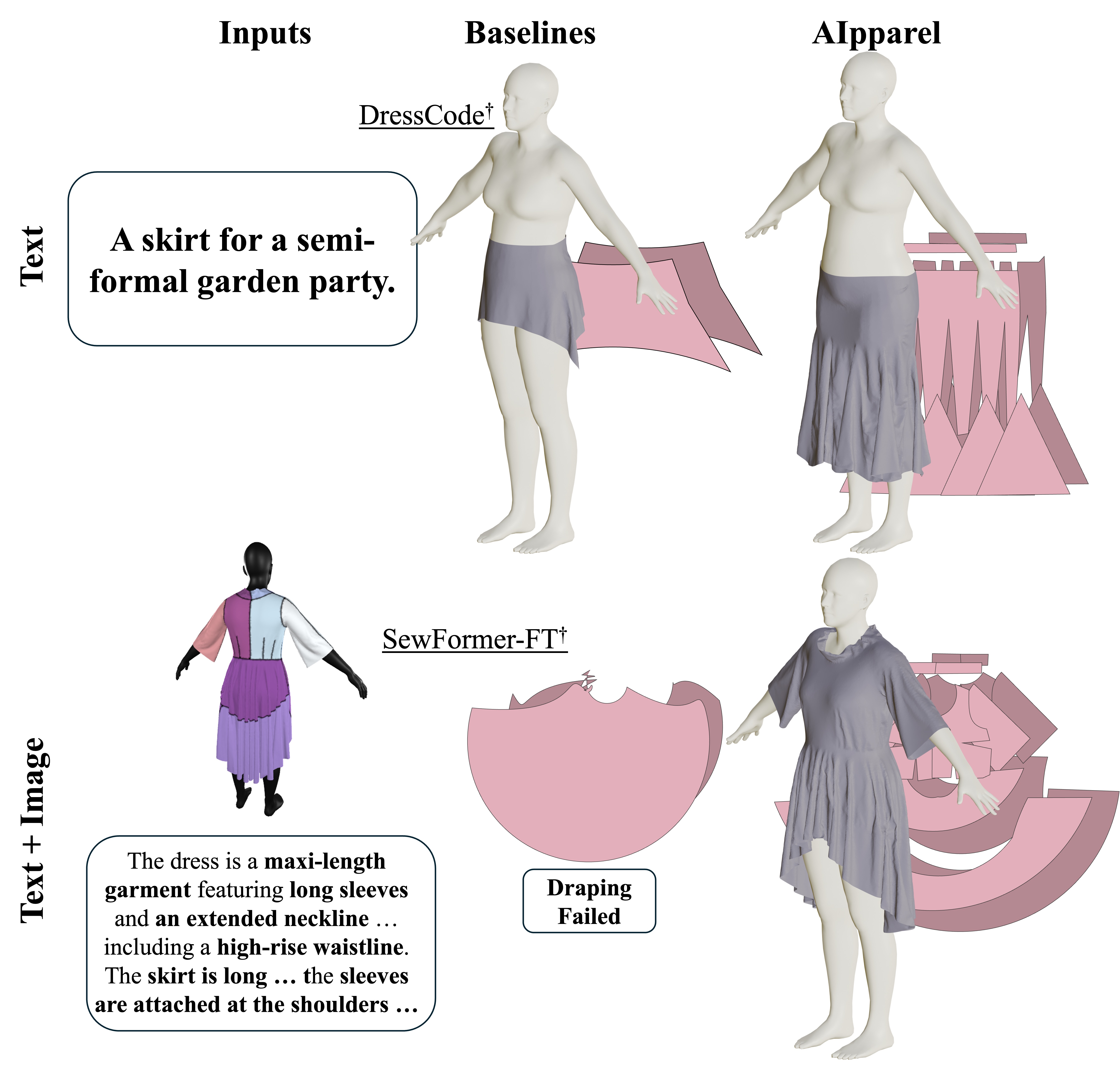}
    \vspace{-2em}
    \caption{\textbf{Multimodal Sewing Pattern Prediction (Qualitative).} AIpparel accurately predicts sewing patterns that follows the inputs better than the baselines. See Sec.~\ref{sec:multi2garment}.}
    \label{fig:multimodal}
\end{figure}

%% file: tables/reasoning_based_pattern_recov.tex
% \begin{table*}[ht]
%     \centering
%     \begin{tabular}{lccccccc}
%     \toprule
%          Method &  Panel L2 $(\downarrow)$ & \#Panel Acc $(\uparrow)$ & \#Edge Acc $(\uparrow)$ & Rot L2 $(\downarrow)$ & Transl L2 $(\downarrow)$ & Stitch Acc $(\uparrow)$ \\\midrule
%     % \midrule
%     SewFormer$^\dagger$ & 22.4 & 45.1 & 22.9 & .15 & 44.3 & 1.93\\
%     DressCode$^\dagger$ & 31.0 & 17.5 & 3.65 & .724 & 114 & 0\\
      
%       \textbf{\methodname}   & \textbf{6.06} & \textbf{81.0} & \textbf{72.9} & \textbf{.0070} & \textbf{2.97} & \textbf{70.6} \\
%       \bottomrule
%     \end{tabular}
%     \vspace{-0.75em}
%     \caption{
%     \textbf{Multimodal Sewing Pattern  Prediction.} 
%     % \todo{100 garments, 20 image, 20 text, 20 both, 20 occasion, 20 editing} \todo{hypothesis: our method is the SOTA in multi-modal garment prediction tasks.} \george{I also put editing as a part of this. We can combine the metrics to make the table tighter.} \todo{Mention that DC and SF cannot work with multimodal and we need to do addition things.}
%     }
%     \label{tab:reason_recov}
% \end{table*}

\begin{table}[t!]
    \centering
    \begin{tabular}{lccccccc}
    \toprule
         Method &  Accuracy $(\uparrow)$ & Panel L2 $(\downarrow)$ \\\midrule
    % \midrule
    SewFormer-FT$^\dagger$ & 10.3 & 22.4 \\
    DressCode$^\dagger$ &  0.6 & 31.0\\
      \textbf{\methodname}  & \textbf{59.0} & \textbf{6.1}  \\
      \bottomrule
    \end{tabular}
    \vspace{-0.75em}
    \caption{
    \textbf{Multimodal Sewing Pattern Prediction.} 
    % Accuracy describes the total accuracy of the predicted sewing patterns compared with the reference. 
    Compared to single-modal methods augmented with existing LMMs, our model outperforms both baselines by a large margin. 
    % thanks to the knowledge we inherent from LLaVA via finetuning.
    % \todo{100 garments, 20 image, 20 text, 20 both, 20 occasion, 20 editing} \todo{hypothesis: our method is the SOTA in multi-modal garment prediction tasks.} \george{I also put editing as a part of this. We can combine the metrics to make the table tighter.} \todo{Mention that DC and SF cannot work with multimodal and we need to do addition things.}
    }
    \label{tab:reason_recov}
\end{table}

%% file: figures/editing.tex
\begin{figure}[t]
    \includegraphics[width=\linewidth]{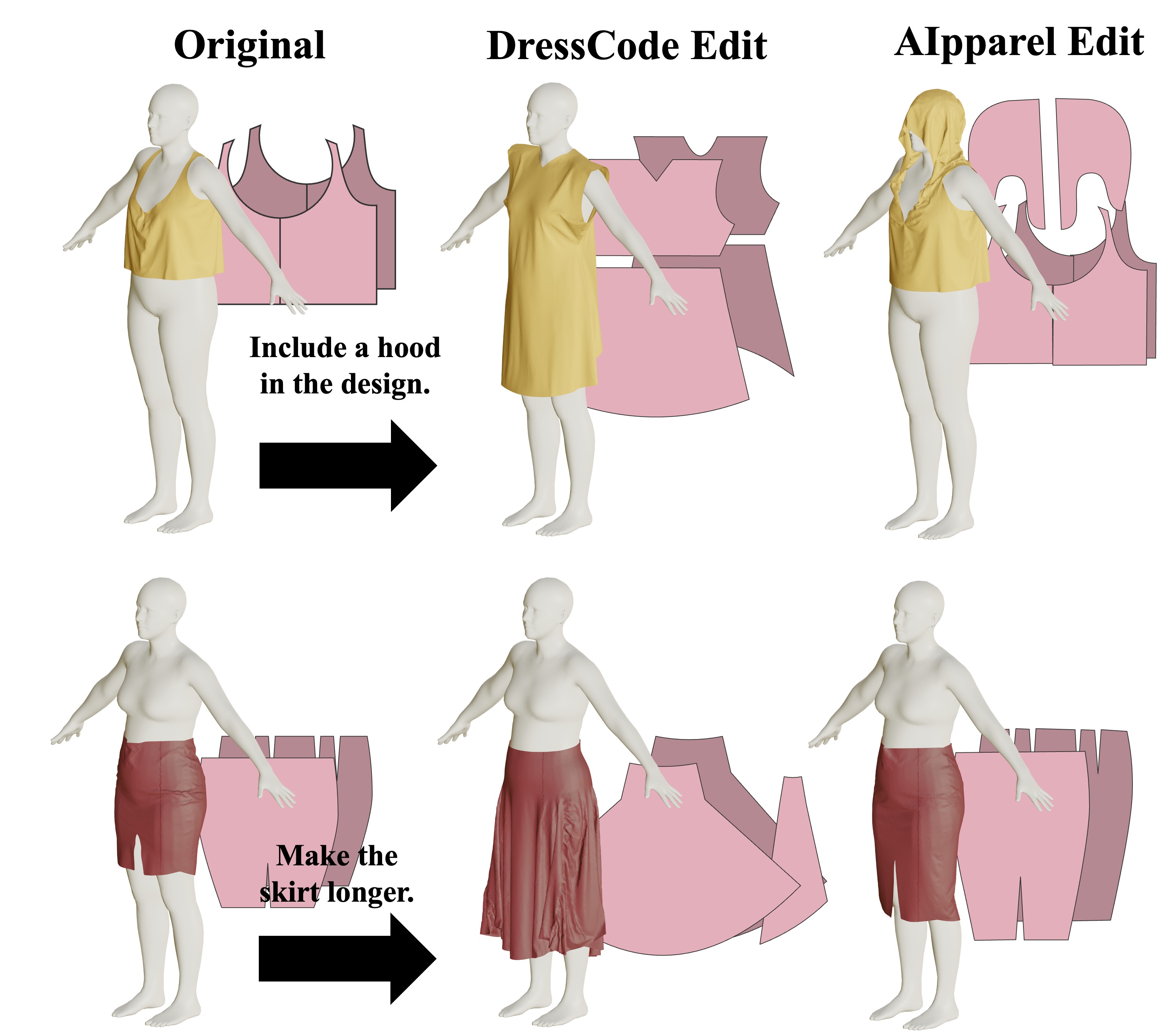}
    \vspace{-0.5em}
    \caption{
        \textbf{Sewing Pattern Editing (Qualitative).}
        % \TODO{Describe ``how'' is this better.}
        Our model follows the editing instructions more accurately compared with the baseline by accurately including a hood to the tank top (top row) and elongating the skirt (bottom row). See Sec.~\ref{sec:garmentedit}.
        % and better preserves the key features of the input sewing patterns.
    }
    \label{fig:editing}
\end{figure}

%% file: tables/editing.tex
% \begin{table}[ht]
%     \centering
%     \begin{tabular}{l@{}cccc@{}}
%     \toprule
%     %%% KEEP THE SAME PRECISIOn %%%%
%     Method &  Accuracy ($\uparrow$)& Shape L2 ($\downarrow$) & Rigid L2 ($\downarrow$)\\
%     % & Stitch Acc. ($\uparrow$)\\
%     \midrule
%     % & \todo{0.0} \\
%     % DC pretrained model on GCdataset
%     Sewformer* & 9.5 & 18.6 & 16.2 \\
%     DressCode* & 37.0 & 13.7 & 113 \\
%     % & 1.39  \\
%     % DC pretrained model on GCdataset
%     % SewFormer (+ iP2P) & 19.4588 & 0.1857 & 0.2496 & 0.0858 & 15.3671 & 0.0139\\
%     % DressCode (+ GPT4V) & nan & 0.0 & nan & nan & nan & 0.0 \\
%       % GPT4V \\
%     % \midrule 
%     \methodname& \textbf{83.4} & \textbf{1.5} & \textbf{1.3} \\
%     % & \textbf{95.1} \\
%     \bottomrule
%     %      &  Panel L2 & \#Panel Acc & \#Edge Acc  & Stitch Acc\\\midrule
%     %   % LLaVA   & \\
%     %   % GPT4V   & \\
%     %   Instruct-P2P + Sewformer   & \\
%     %   LLaMa + DressCode   & \\
%     %   \methodname   & 
%     \end{tabular}
%     \caption{\textbf{Language-conditioned sewing pattern Editing}\todo{Make a comment on we've already done the best effort.}}
%     \label{tab:editing}
% \end{table}

\begin{table}[t]
    \centering
    \begin{tabular}{l@{}cccc@{}}
    \toprule
    %%% KEEP THE SAME PRECISIOn %%%%
    Method &  Accuracy ($\uparrow$)& Panel L2 ($\downarrow$)\\
    % & Stitch Acc. ($\uparrow$)\\
    \midrule
    % & \todo{0.0} \\
    % DC pretrained model on GCdataset
    Sewformer* & 9.5 & 18.6 \\
    DressCode* & 37.0 & 13.7 \\
    % & 1.39  \\
    % DC pretrained model on GCdataset
    % SewFormer (+ iP2P) & 19.4588 & 0.1857 & 0.2496 & 0.0858 & 15.3671 & 0.0139\\
    % DressCode (+ GPT4V) & nan & 0.0 & nan & nan & nan & 0.0 \\
      % GPT4V \\
    % \midrule 
    \methodname& \textbf{83.4} & \textbf{1.5} \\
    % & \textbf{95.1} \\
    \bottomrule
    %      &  Panel L2 & \#Panel Acc & \#Edge Acc  & Stitch Acc\\\midrule
    %   % LLaVA   & \\
    %   % GPT4V   & \\
    %   Instruct-P2P + Sewformer   & \\
    %   LLaMa + DressCode   & \\
    %   \methodname   & 
    \end{tabular}
    \vspace{-0.75em}
    \caption{
        \textbf{Sewing Pattern Editing.}
        We use SOTA LMMs such as GPT-4V and DALL-E to facilitate both baselines to perform this multimodal editing task.
        Our model still outperforms both baselines by a large margin.
    % \todo{Make a comment on we've already done the best effort.}
    }
    \label{tab:editing}
\end{table}

%% file: tables/ablation.tex
\begin{table}[t]
    \centering
    \begin{tabular}{l@{}cccc}
    \toprule
         Methods &  Accuracy ($\uparrow$) & Panel L2 ($\downarrow$) & Time ($\downarrow$)\\
         \midrule
        DressCode & 38.4 & 22.4 & 52.2s \\
        % \methodname - pretraining\\
        Ours w.o. reg. & 79.0 & 7.2 & 3.4s\\
        \textbf{Ours} &   \textbf{85.0} & \textbf{6.1} & \textbf{2.1s}\\
    \bottomrule
    \end{tabular}
    \vspace{-0.75em}
    \caption{
    \textbf{Ablation.} 
    % We show text-to-garment metrics using the NeuralTailor dataset following DressCode's experiment setup.
    Our tokenizer outperforms DressCode in all metrics while being more than 25 times faster at inference time.
    Our objective (Eq.~\ref{eq:obj}) also improves performance compared to the cross-entropy-only variant. 
    % \todo{NeuralTailor data, pretrained DressCode, text-to-garment reconstruction. GPT train from scratch.} \todo{Show our data too?}
    % \vspace{-0.7cm}
    }
    \label{tab:ablation}
\end{table}

% TABLE: 
% % - Our ds + LLAVA + DressCode tokenizer
% % - Our ds + LLAVA + Our tokenizer
% % - NeuralTailor + LLAVA + DressCode tokenizer
% % - NeuralTailor + LLAVA + Our tokenizer

% % - Our ds + GPT2 + DressCode tokenizer (=0, failed :|)
% % - Our ds + GPT2 + Our tokenizer      
% Step 1: Process garment from NeuralTailor (1. panel starting point; 2. pad panel; 3. quantize...) 
% - NeuralTailor + GPT2 + DressCode tokenizer (pretrained done)
% - (** PRIORITIZE **) NeuralTailor + GPT2 + Our tokenizer 
% (deprioritize) - NeuralTailor + GPT2 pretrained + DressCode tokenizer
% (deprioritize) - NeuralTailor + GPT2 pretrained + Our tokenizer

%% file: figures/ablation.tex
\begin{figure}[ht]
    \includegraphics[width=\linewidth]{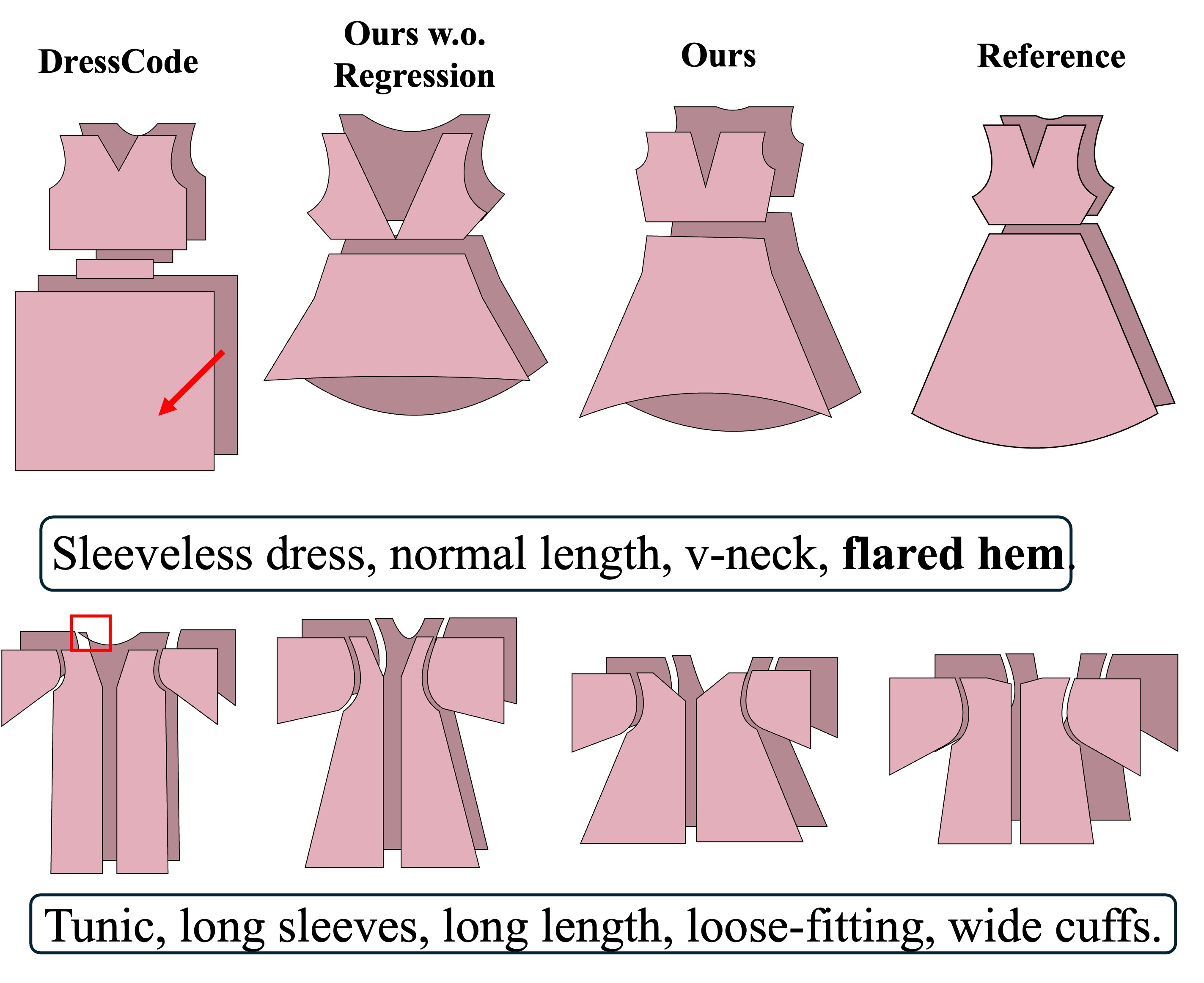}
    \vspace{-1em}
    \caption{
    \textbf{Ablation (Qualitative).} 
      DressCode's tokenizer produces unrealistic patterns (second row, boxed region) and does not match the text input (i.e., ``flared hem"). In contrast, our tokenizer outputs geometrically regular sewing patterns accurately aligning with the inputs. See Sec.~\ref{sec:ablation}.  \vspace{-2em}
    }
    \label{fig:ablation}
\end{figure}

%% file: sec/5_conclusion.tex
\section{Discussion}
\label{sec:discussion}
We introduce \methodname, a 7B-parameter multimodal foundation model for garment sewing patterns. 
To train \methodname, we curate GCD-MM, a large-scale dataset with complex sewing patterns and multimodal annotations. 
Moreover, we develop a novel sewing pattern tokenizer and a mixed training objective for fine-tuning LMMs on GCD-MM. 
\methodname achieves state-of-the-art results on single-modal and multimodal sewing-pattern-generation tasks, enabling new applications like language-driven sewing pattern editing. 

% \paragraph{Limitations and Future Directions.} 
% please refer to the supplementary material.
% \paragraph{Societal Impact}
% \todo{Disscuss social impact}
% \paragraph{Conclusion.} 
% Supplementary has more discussion on limitations and future works.
% \textcolor{red}{need 1--3 sentences ending the paper on a big note}

% \vspace{-1em}
\noindent \textbf{Limitations and Future Work.}
%Our work is not without limitations. 
While the current representation enables the digitalization of complex sewing patterns, it is still constrained to garments representable by manifold surfaces. 
Design elements like pockets require non-manifold structures. 
A promising direction is to develop an efficient representation that accurately models non-manifold features while remaining compatible with LMMs. 
Fabricating the generated garments is another interesting direction, which requires consideration of physical and material constraints during sewing pattern prediction.

% \vspace{-1em}
\noindent \textbf{Broader Impacts.} 
While we believe our model can advance AI-assisted fashion design, we acknowledge potential risks we inherit from the pre-trained LLaVA model. 
For instance, generative AIs can spread misinformation or create biases potentially harmful to society. We do not condone these and other improper usage of our model.

% \vspace{-1em}
\noindent \textbf{Conclusion.}
%\methodname is the first work that enables sewing pattern generation with image-text-garment input. Compared to previous works, \methodname moves one step closer to building an AI virtual tailor that intelligently assists fashion designers in their daily designing cycle. Moreover, our proposed sewing pattern tokenization scheme also offers new capability and efficiency in encoding sewing patterns as machine learning-friendly representations. 
%
Vision-language and other large multimodal models capture web knowledge and enable reasoning for many downstream applications. By fine-tuning LMMs to understand sewing patterns, we take first steps towards a vision-language-garment model that transfers web knowledge to garment generation and editing, unlocking a plethora of applications for fashion design and fabrication.

%% file: sec/6_ack.tex
\section{Acknowledgements}
\label{sec:ack}
Google, an ERC Consolidator Grant No. 101003104 (MYCLOTH), an ARL grant W911NF-21-2-0104, a Vannevar Bush Faculty Fellowship, and LVMH support the project. Finally, we thank Stanford's Marlowe cluster for providing GPU computing for model training and evaluation. 

%% file: sec/X_suppl.tex
{\LARGE\bfseries Supplementary Materials}
\makeatletter
\renewcommand{\l@section}{\@dottedtocline{1}{1.5em}{2em}}
\renewcommand{\l@subsection}{\@dottedtocline{2}{4.0em}{2.8em}}
\renewcommand{\l@subsubsection}{\@dottedtocline{3}{7.4em}{4.5em}}
\makeatother
\tableofcontents
\changelocaltocdepth{3}
\SupplementaryMaterials
% \renewcommand\labelenumii{\theenumi\arabic{enumii}.}
% \begin{enumerate}[label={S\arabic*.}]
%     \item Details on GarmentCodeData-Multimodal (GCD-MM)
%     \begin{enumerate}
%         \item Text Description Generation
%         \item Generation of Editing Data Sample
%     \end{enumerate}
% \end{enumerate}
% \input{figures/supp_editing/fig}

% \newpage

% \input{figures/supp_image2garment/fig}

% \newpage

% \input{figures/supp_text2garment/fig}

% \newpage

% \input{figures/supp_sewformer_comp/fig}

% \newpage

% \input{figures/supp_mm/fig-occasion}

% \newpage

% \input{figures/supp_mm/fig-gi}

% \newpage
% \input{figures/supp_mm/fig-gi2}

% \section{Video Demo}
% We attach a video showcasing garment simulation results using garments draped with predicted sewing patterns from AIpparel from multimodal inputs. Notice that our model's prediction is directly usable in traditional simulation pipelines and enables creating animation with different character movements and texture editing \todo{fill in details once the video is there.}

\section{Details on GarmentCodeData-Multimodal (GCD-MM)}
\label{sec:dataset_detail}
We expand on the data curation process of GCD-MM. Specifically, in Sec.~\ref{sec:text_detail}, we detail the specifics of generating text descriptions for the sewing patterns. In Sec.~\ref{sec:editing_detail}, we elaborate on how we generate the edited sewing patterns and their associated editing instructions. Lastly, in Sec.~\ref{sec:sewing_pattern_stats}, we show statistics comparing sewing patterns in GCD-MM and sewing patterns in previous datasets used by DressCode~\cite{he2024dresscodeautoregressivelysewinggenerating} and SewFormer~\cite{liu2023sewformer}.

\subsection{Text Description Generation}
\label{sec:text_detail}
We generate two types of sewing pattern descriptions for each garment in GCD-MM. The first is a detailed natural language description of the sewing pattern, while the second outlines a suitable occasion for wearing the garment. For captioning purposes, we use a standardized body type corresponding to the mean shape and pose derived from SMPL.

Obtaining pattern descriptions happens in two steps. First, we generate keywords describing the simulated garments using the design parametrization of each garment. 
Generated based on GarmentCode \cite{korosteleva2023garmentcode}, each garment is characterized by a set of continuous and categorical parameters. 
% These form part of the GCD dataset. 
% Our rule-based description generation tackles each of the parameters differently: 
We generate descriptions for each garment using the following rules:
\begin{itemize}

    \item \textbf{Categorical parameters}: We assign the categorical label when appropriate. For instance, a \verb|godet skirt| is classified as such. Some categorical parameters do not suffice - a \verb|shirt| can signify anything from a crop top to a dress. For these instances, we add additional checks consulting additional parameters. 

    \item \textbf{Continuous parameters}: We define thresholds and assign different qualitative labels for garments above and below them. Parameters such as \verb|sleeve length| or \verb|collar width| are obvious examples. 

    \item \textbf{Dependent parameters}: Most parameters have no impact on the final garment, as they only become relevant when certain categorical parameters are set. We design rules that consider these edge cases. Only when a \verb|godet skirt| is set, does the \verb|num inserts| become relevant. We include all relevant dependent parameters that have a structural effect on the garment. 
\end{itemize}

\begin{figure}
    \centering
    \includegraphics[width=\linewidth]{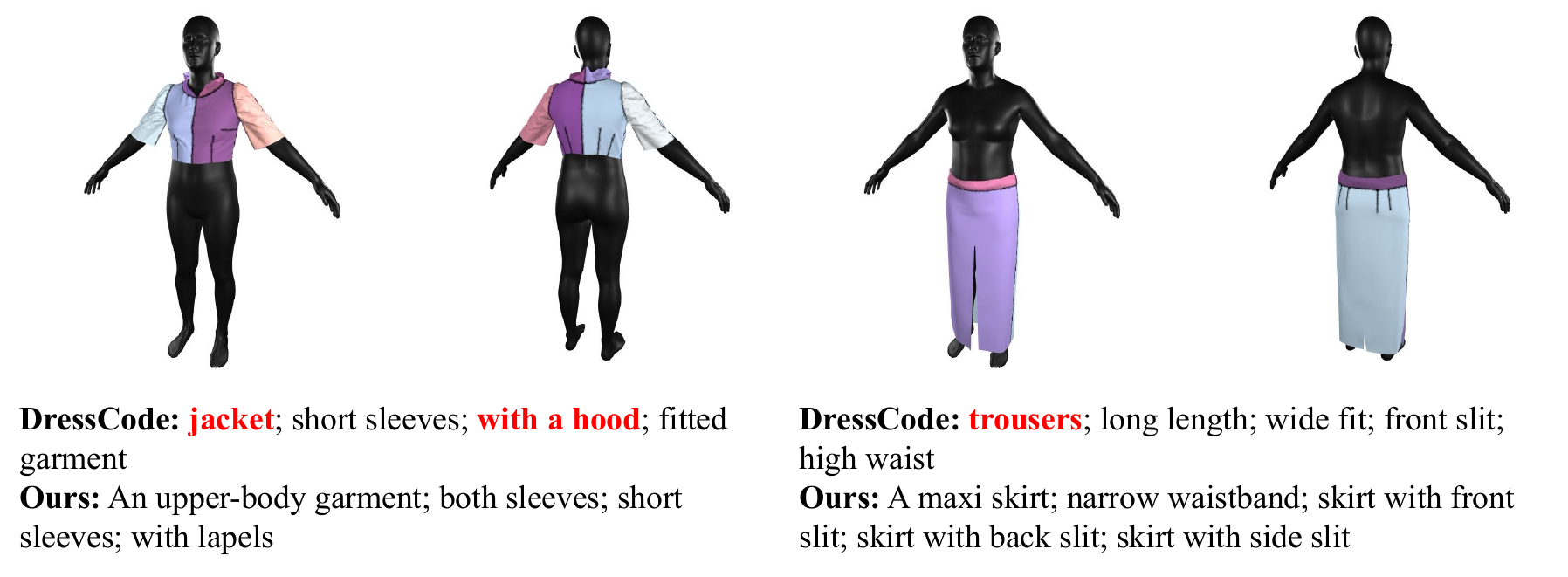}
    \caption{\textbf{Comparison between our Short Captions and DressCodes'.} This figure shows the short captions created by DressCode and our method for two different garments. DressCode produces keywords that do not align with the garment (red).}
    \label{fig:data_comparison}
\end{figure}

Similar to DressCode, we first generate a garment type description and a collection of keywords that contain the specific description based on our rule-based approach. 
Note that each rule can contribute several keywords. See Figure~\ref{fig:data_comparison} for the examples. 

In the second stage, we use these generated keywords in combination with a render of the front and back of the garment to prompt GPT-4o. 
We construct the prompt such that GPT-4o objectively describe the garment using the characteristic features of the garment provided by the generated keywords and renders.
In addition, we include instructions to focus on information crucial for our learning problems, such as panel connectivity and stitching patterns, while ignoring irrelevant information, such as colors or interpretations.

The following is the system prompt that we used:

\begin{tcolorbox}[colback=promptbg, colframe=promptborder, boxrule=0.5mm, arc=2mm, left=2mm, right=2mm]
You are a fashion expert tasked with providing concise and neutral descriptions of garments based on the provided textual information. Your descriptions should focus on specific stitching details and how different panels are connected (such as seam placements and stitching patterns), as well as any distinctive characteristics and design elements of the garment. When describing the garment's appearance, use precise and concrete language, avoiding generic phrases or broad descriptions. Do not mention that seams are visible; instead, describe where seams or panels are located to indicate construction details. Do not include any impressions, subjective interpretations, or unobservable aspects. Avoid mentioning colors or any references to images. Keep the descriptions brief and to the point, avoiding unnecessary words. Use only the information provided.
\end{tcolorbox}

Here we present the user prompt:
\begin{tcolorbox}[colback=promptbg, colframe=promptborder, boxrule=0.5mm, arc=2mm, left=2mm, right=2mm]
Please generate a concise and neutral description of a garment, focusing on specific stitching details, how different panels are connected, and including any distinctive characteristics and design elements, based on the following information:

- **Title**: \{title\}

- **Description**: \{description\}

Provide a brief description that emphasizes stitching and construction details (such as seam placements, panel connections, and stitching patterns), along with precise visual observations about the garment's appearance, including style, silhouette, length, and any unique design features or distinctive characteristics. Avoid using generic phrases or broad descriptions; instead, provide specific details about the garment's features. Do not mention that seams are visible; instead, describe where seams or panels are located to indicate construction details. Do not include any impressions, subjective interpretations, or unobservable aspects. Avoid mentioning colors or any references to images. Keep the description succinct and avoid unnecessary words. Use only the information provided.
\end{tcolorbox}

The second type of caption describes an occasion for which a garment is suitable. In this prompt, we ask the model not to pay attention to the garment's colors which only highlight different panels and are not semantically relevant. Instead, we ask it to focus on the shape and description. We use the same information as before to prompt GPT-4o.
This is the system prompt:
\begin{tcolorbox}[colback=promptbg, colframe=promptborder, boxrule=0.5mm, arc=2mm, left=2mm, right=2mm]
You are an expert in fashion design and garment analysis. When provided with images of garments and their metadata, focus solely on their shape and stitching. Note that different colors in the images represent different panels of the garment and are not indicative of style or color choices. Ignore colors and any visible seams meant only for stitching information. The metadata includes a title and a description, which is a list of short attributes; use these to inform your understanding. Based on this information, provide only a detailed, but concise, description of a single occasion where the given garment would be appropriate to wear. Do not include any other information in your response.
\end{tcolorbox}

and here is the user prompt:
\begin{tcolorbox}[colback=promptbg, colframe=promptborder, boxrule=0.5mm, arc=2mm, left=2mm, right=2mm]
Given the following garment's metadata and images (remember that colors and seams are only for panel representation and stitching information), please provide only a detailed, but concise, description of a single occasion where this garment can be worn. Do not include any other information in your response.

Here is the metadata:

Title: \{title\}

Description: \{description\}
\end{tcolorbox}

\paragraph{Effect of GPT version in caption quality.}
\begin{figure}
    \centering
    \includegraphics[width=\linewidth]{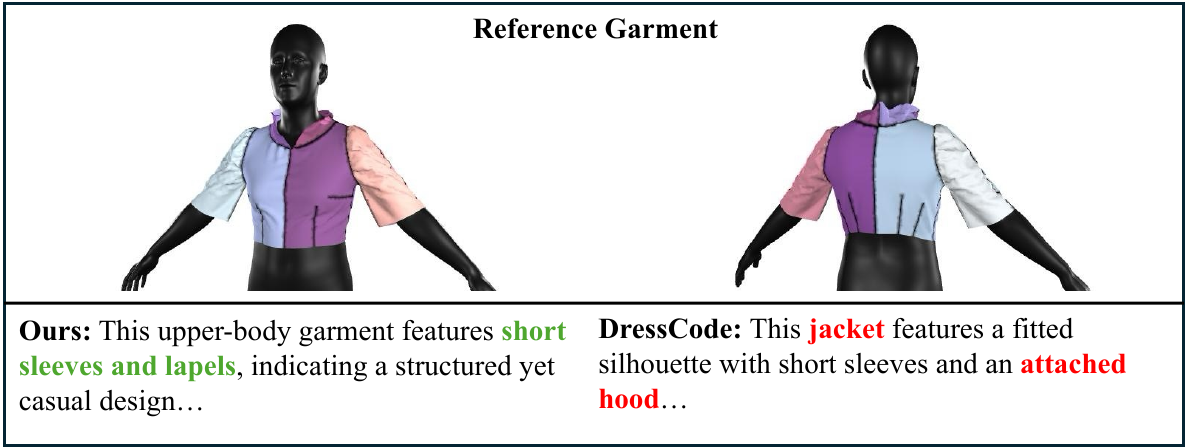}
    \caption{\textbf{Captions generated using GPT-4V.}}
    \label{fig:garment_retexturing}
\end{figure}
While GPT-4o potentially increases the accuracy of generated captions, the in-context knowledge about various design parameters crucially helps the model to generate captions more faithful to the garment design. Fig.~\ref{fig:garment_retexturing} shows the same captions in Fig.~\ref{fig:data_comparison} of supp, generated instead using GPT-4V. Notice that DressCode's caption contains severe flaws (in red) due to inaccurate in-context prompting. Ours do not have these flaws because we prompt GPT-4V with design-parameter-inspired content. We will update Fig.S1 to include this example in the revision. 

\subsection{Generation of Editing Data Sample}
\label{sec:editing_detail}
% \todo{todo}

To generate paired garments representing before-and-after edits, we use design parameters from the GCD dataset and systematically apply one of five pre-defined transformation rules. 
The modified design parameters are then converted into garments using GarmentCode \cite{korosteleva2023garmentcode}.

Each garment from GCD is first evaluated to determine which transformation rules are applicable. 
One rule is then randomly selected and applied. 
Due to limitations in GarmentCode's design space, not all edited design parameters can be converted into sewing patterns. 
As a result, GCD-MM comprises 120k garment pairs that are successfully generated from the 130k garments in GCD, while approximately 10k garments remain unpaired.

The transformation rules include adjustments to garment lengths (sleeves, pants, skirts), collar type changes, modifications to garment symmetry, toggling the presence of hoods, and structural edits to style elements (e.g., changing the number of inserts in godet skirts). 
Each rule takes the existing design parameters as input and applies a targeted change. 
For instance, length adjustments alter sleeves, pants, or skirts by 50\% of their initial length, constrained by the maximum length specified in GarmentCode. 
Similarly, collar types are randomly reassigned from a predefined set, garment symmetry is toggled, and hoods are added or removed.

These rules are designed for three key reasons: (1) they produce clear and concise edits that can be succinctly summarized; (2) they encompass varying levels of editing complexity, from minor panel length adjustments to major structural modifications involving new panels and altered stitching; and (3) for all garments in the dataset, at least one rule can always be applied.

To document each transformation, we generate descriptive sentences for the edited garments using a rule-based approach. Here are a set of examples:

\begin{tcolorbox}[colback=promptbg, colframe=promptborder, boxrule=0.5mm, arc=2mm, left=2mm, right=2mm]
Godet skirt: \hspace{0.5cm} \textit{"Increase the number of inserts in the skirt by \$x."} \newline
Pants: \hspace{1.4cm} \textit{"Make the pants longer."}  \newline
Shirt: \hspace{1.5cm} \textit{"Switch the collar type from \$currCollar to \$newCollar."} 
\end{tcolorbox}

In total, the defined rules enable 52 distinct, describable modifications, ensuring a diverse and well-documented dataset of garment editing pairs.

\subsection{Sewing Pattern Statistics}
% \subsection{Sewing Pattern Statistics -- TO REMOVE} - gy:I think we should keep it.
\begin{figure}[t!]
    \centering
    \includegraphics[width=\linewidth]{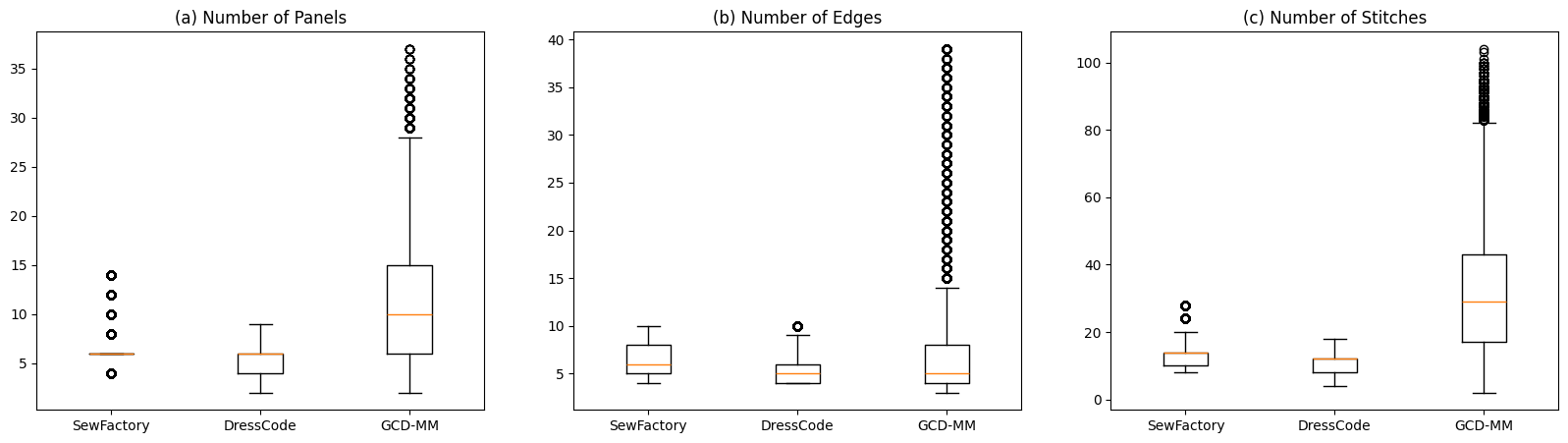}
    \caption{\textbf{Dataset Statistics Comparisons.} Notice that GCD-MM in general contains larger variations in the number of panels, edges, and stitches in the sewing patterns. This poses additional challenges in designing a sewing pattern generation method with GCD-MM.}
    \label{fig:dataset_stats}
\end{figure}
\input{tables/dataset_stats}
\label{sec:sewing_pattern_stats}
GCD-MM uses sewing patterns fitted on a default body from the GarmentCodeData (GCD) dataset~\cite{korosteleva2024garmentcodedatadataset3dmadetomeasure}, which are procedurally generated sewing patterns using the programming abstraction of GarmentCode~\cite{korosteleva2023garmentcode}. Compared with the sewing patterns used by SewFormer~\cite{liu2023sewformer} and DressCode~\cite{he2024dresscodeautoregressivelysewinggenerating}, GCD contains more complicated and diverse sewing patterns. For detailed documentation and comparison with existing datasets and procedural sewing pattern generators, please refer to GarmentCodeData~\cite{korosteleva2024garmentcodedatadataset3dmadetomeasure}. Here, we briefly show some statistics comparing these different datasets. 

GCD exhibits more diverse and detailed garment feature variations than the previous dataset, including fitted garments, correct sleeve shapes, more collar types, more skirt types, cuffs, and asymmetric features (tops, asymmetric skirt cuts). All of these characteristics make sewing patterns from GCD more complicated than existing sewing pattern datasets. 

\begin{figure}[t!]
    \centering
    \includegraphics[width=\linewidth]{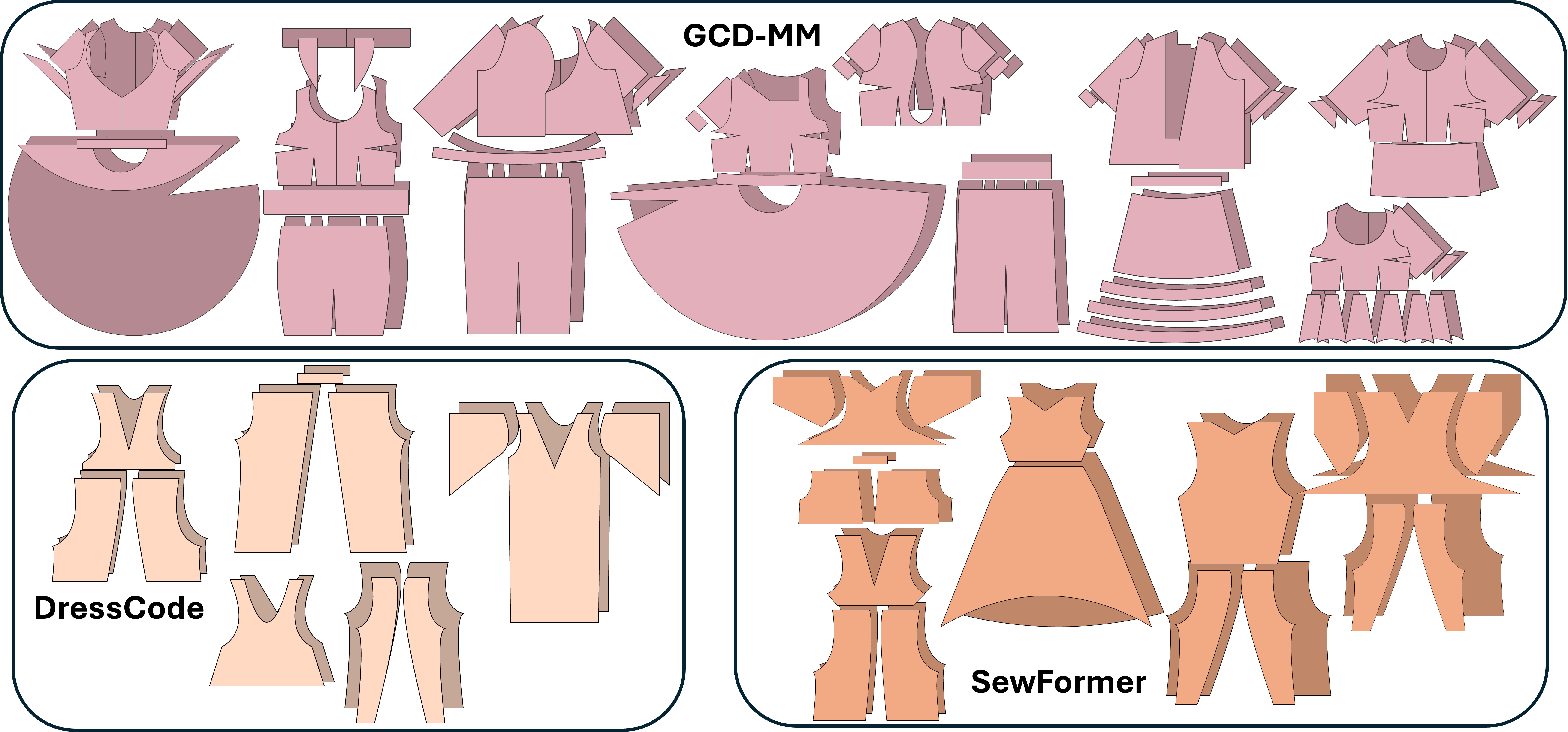}
    \caption{\textbf{Visualization of Sewing Patterns}. Random sewing pattern samples from the datasets used by \methodname and the baselines are visualized. Notice that compared to prior works, GCD-MM exhibits more complex sewing patterns in general.}
    \label{fig:dataset_vis}
\end{figure}

Comparatively, datasets used by SewFormer~\cite{liu2023sewformer} and DressCode~\cite{he2024dresscodeautoregressivelysewinggenerating} are procedurally generated sewing patterns from an older programming abstraction~\cite{KorostelevaGarmentData}. 
While this programming abstraction can also generate sewing patterns for the types of garments described above, all its variations are from changes in the vertex and control point positions while fixing the number of panels, edges, and stitches the same. This constraint significantly limits the variations exhibited in the datasets used by SewFormer and DressCode. \Cref{fig:dataset_vis} showcases randomly sampled sewing patterns as well as their draped renderings from GCD and sewing patterns used by SewFormer and DressCode. We see that sewing patterns from GCD are generally more complex and diverse than the previous dataset. \Cref{tab:dataset_stats} and \Cref{fig:dataset_stats} show a statistical comparison in terms of the number of edges, panels, stitches, and edge types between sewing patterns in GCD-MM, SewFactory~\cite{liu2023sewformer}, and dataset used by DressCode~\cite{he2024dresscodeautoregressivelysewinggenerating}. Notice that comparatively sewing patterns in GCD-MM exhibit the largest variation in all of the statistics, demonstrating the difficulty of the dataset. In particular, because of this difficulty gap, previous methods such as SewFormer and DressCode exhibit poor performance despite fine-tuning their network on GCD-MM. See \Cref{sec:further_results]} for details. 

\section{Implementation Details on \methodname}
We include details about the network architecture and training hyperparameters of \methodname.

\subsection{Network Architecture}
As described in \Cref{sec:method} of the main paper, \methodname is built on top of LLaVA-1.5 7B~\cite{liu2023llava}. Therefore, the majority of the network, except for the newly added regression heads $g^{(\text{e})}_\theta, g^{(R)}_{\theta^\prime}$, and the positional embedding projection layers $h^{(\text{e})}_\varphi, h^{(R)}_{\varphi^\prime}$, are identical to LLaVA-1.5 7B. 
For completeness, we only summarize the key parameter values we used here. 
Please refer to their paper for architectural details. LLaVA-1.5 7B fine-tunes LLama 2~\cite{touvron2023llama2openfoundation} with a vision encoder on a visual question-answer dataset. Specifically, it has a context length of 4049 and a hidden dimension of 4096. Its language model is a 32-layer transformer with 32 head attention layers. Its vision encoder is CLIP~\cite{radford2021learningtransferablevisualmodels}. Each image is converted into 255 clip tokens before getting projected into the language model's embedding space using a custom projector. 

To extend LLaVA-1.5 7B for sewing pattern prediction, we expand the vocabulary of the model to include the special tokens defined in \Cref{sec:training} of the main paper. In total, this results in 122 additional tokens added to the vocabulary of LLaVA-1.5 7B. Each of the tokens is initialized to be the average embedding from the existing vocabulary.

Besides additional vocabulary, we also add two additional regression heads $g^{(\text{e})}_\theta, g^{(R)}_{\theta^\prime}$, and the positional embedding projection layers $h^{(\text{e})}_\varphi, h^{(R)}_{\varphi^\prime}$ to the architecture described above. 
As described in \Cref{sec:method} of the main paper, the regression heads will take the output hidden embedding from the language transformer to regress vertex and control point positions using $g^{(\text{e})}_\theta$ and the transformation with $g^{(R)}_\theta$. Specifically, both of the regression heads are two-layer perceptrons with ReLU non-linearity. 
Both heads map the 4096-dimensional output embedding to the parameter space. For  $g^{(\text{e})}_\theta$, the output dimension is $8$, representing vertex and control points in different channels. Specifically, the first two channels as vertex regression, mapping to the second endpoint of the associated edge. The next four are used for control points to the quadratic and cubic Bézier curves. Finally, if the associated edge is an arc, the last two channels are used to map to an additional point on the arc besides the two endpoints. During training, only the associated channels for each edge are supervised and the unused channels are masked out for back propagation. With the same architecture, $g^{(R)}_\theta$ has an output dimension of 7, with the first 3 being the translation and the last four being the rotation represented in quaternion. 

Finally, the positional embedding layers are also two-layer perceptions with ReLU non-linearity. $h^{(\text{e})}_\varphi$ maps the 2-dimensional vertex coordinate to a 4096-dimensional hidden embedding. The output is then added to that edge type token's vocabulary embedding before inputting through the language transformer. Similarly, $h^{(R)}_\varphi$ maps the 7-dimensional transformation for each panel to a 4096-dimensional hidden embedding. Then the output is added to the vocabulary embedding of the transformation token \texttt{<R>}. 

Both the regression heads and the positional embedding projection layers are initialized to have zero weights in the final layer so that the output before fine-tuning is unaltered.

\subsection{Training Details}
\label{sec:supp_training_detal}
\methodname is trained for a total of 12,750 steps with a total batch size of 320, and a learning rate of $0.00005$ with cosine learning rate decay to zero in 15,000 steps. We also warm-start the fine-tuning from zero learning rate to the default in the first 100 steps. We use $\lambda=0.1$ to balance the regression losses and the cross-entropy loss in \Cref{eq:obj} of the main paper.  
We use DeepSpeed ZeRO Stage 2~\cite{deepspeed} to parallelize the training on 8$\times$H100 GPUs. 
The entire training took around 312 H100 GPU hours. We train on all modalities in our GCD-MM jointly. Specifically, we include four different modalities from GCD-MM: \textit{text $\rightarrow$ sewing pattern, image $\rightarrow$ sewing pattern, text and image $\rightarrow$ sewing pattern}, and \textit{sewing pattern and editing instruction  $\rightarrow$ edited sewing pattern}. During each training step, the batch is formed by randomly sampling each of the four modalities with a preset sampling ratio. Specifically, we sample images, texts, image+text, and editing data with the ratio of 3:2:4:1. We randomly split our dataset into 90\%, 5\%, and 5\% for training, validation, and testing. All of our qualitative results are samples from the testing split. While previous works~\cite{liu2023sewformer, he2024dresscodeautoregressivelysewinggenerating, NeuralTailor2022} use relative coordinates to represent the control point coordinate, we use absolute coordinates to represent the additional edge parameters. Prior to training, we normalize vertex coordinates and transformation using the global mean and standard deviation computed from all sewing patterns in GCD-MM. Additionally, for input to the positional embedding projection layers, we discretize the input into 256 discrete values ranging between $\pm4$ standard deviation values for robustness during generation.  

\section{Experiment Details And Additional Results}
\label{sec:further_results]}
We detail the experiment setup and baselines for the result section (\Cref{sec:experiments} of the main paper). Further, we also include additional ablation results and qualitative comparisons. 

\subsection{Sewing Pattern Prediction from Images}
\paragraph{Setup \& Baseline Details.}
We will describe the image-to-garment prediction experiment showcased in \Cref{sec:img2garment} of the main paper in detail. 
We will also report comparisons on two datasets: GCD-MM  and SewFactory. 

For GCD-MM, we use our model trained with multimodal data described in \Cref{sec:supp_training_detal} to evaluate the qualitative and quantitative results showcased in \Cref{tab:image2garment} and \Cref{fig:image2garment} of the main paper. To compare with SewFormer~\cite{liu2023sewformer}, we adapt its pre-trained model for sewing pattern prediction on GCD-MM. 
Specifically, we expand the per-panel query embedding from its default number of 23 to 75 to accommodate all the different panel classes present in GCD-MM. We initialize the newly added panel query embeddings as the average embedding from the pre-trained weights. Similarly, we expand the per-edge embedding from 14 to 39. Furthermore, because GCD-MM contains cubic Bézier curves and arcs, which the SewFactory dataset does not have, we also extend per-edge parameterization from using four channels (2+2: endpoint + optional quadratic Bézier control points) to seven channels (2+4+1: endpoint, control point parameters, arc flag). 
Specifically, the arc flag takes a value of 0 or 1, indicating if the edge is an arc. If the arc flag is 1, the first two control points would take a value equal to the relative coordinate of the third point on the arc. If the arc flag is zero, then the four channels will be the relative coordinates of the two control points in the Bézier curve. We keep the network architecture the same except for the above modifications. We fine-tune the pre-trained SewFormer model adapted as above for a total of 16 epochs on the same training split \methodname is trained on, using a learning rate of 0.00005 and a batch size of 8 on 2$\times$Quadro RTX 8000 GPUs. Except for these, we use the default hyperparameters provided by SewFormer. The validation loss no longer increases after 16 epochs, so we stop the training and use it for comparison. 

For comparison on SewFactory, we use the pre-trained SewFormer model as our baseline. However, because the SewFormer authors did not release their train and test split, we show a comparison on a custom test set for this experiment. Specifically, we first train \methodname on SewFactory data, with a different random split, from scratch for a total of 3750 steps on 8$\times$A100 GPUs using the same hyperparameter settings as described in \Cref{sec:supp_training_detal}. Then, we evaluate our model on the custom test set. In this way, we ensure a fair comparison with the baseline as the test set should contain a mixture of training and testing examples for both methods. 

\begin{sidewaysfigure}
    \includegraphics[width=\textwidth]{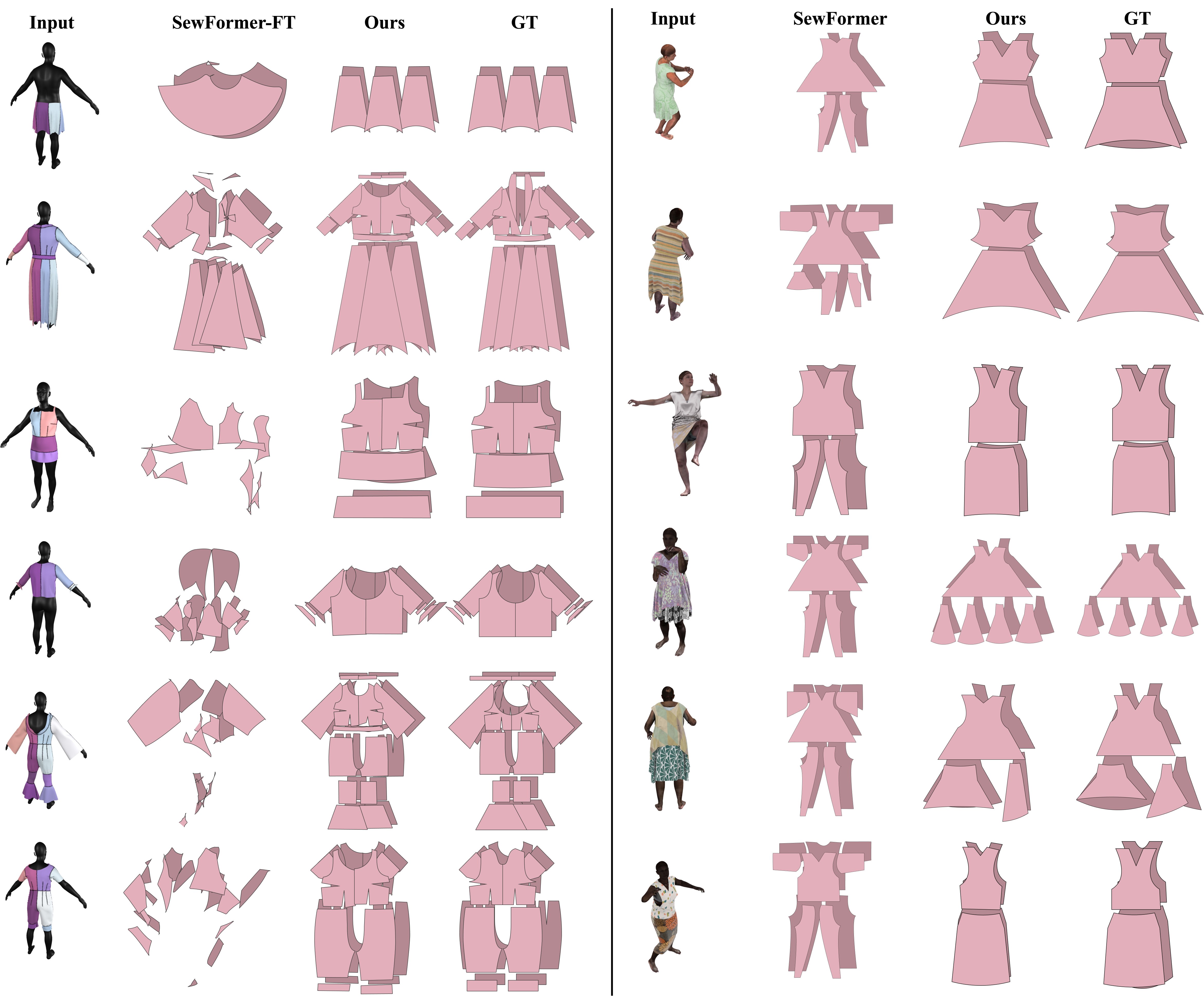}  
    \caption{\textbf{Additional Image to Sewing Pattern Visualizations.}}
    \label{fig:img2garment_supp}
\end{sidewaysfigure}

\paragraph{Additional Qualitative Visualization.}
\Cref{fig:img2garment_supp} showcases additional image-to-garment prediction result comparisons to the SewFormer baseline in both the GCD-MM (left) and SewFactory (right) datasets. Our model in general predicts more correct sewing patterns following the guidance of the input image than SewFormer. 

\begin{figure}
    \centering
    \includegraphics[width=\textwidth]{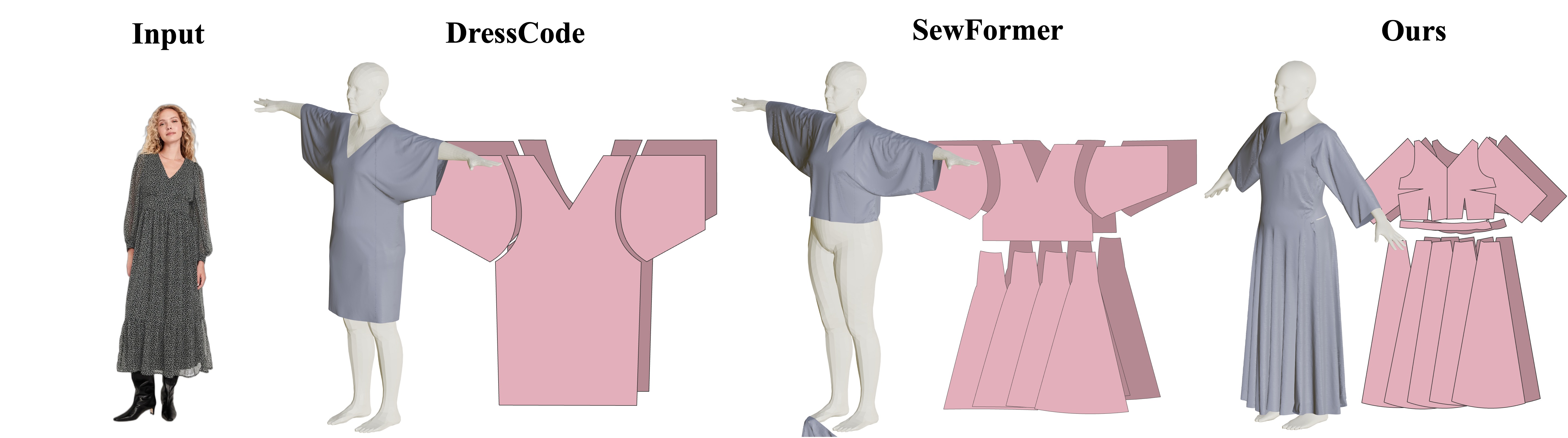}
    \caption{\textbf{In-the-wild Image to Garment Example.} Our model is able to predict a sewing pattern more aligned with the input image compared to the baselines. Notice that SewFormer did not drape correctly, resulting in a missing bottom.}
    \label{fig:reallife_supp}
\end{figure}

\paragraph{Sewing Pattern Prediction from In-the-wild MultiModal Inputs.}
While \methodname is trained on procedurally generated sewing patterns and annotations, it is able to generalize the in-the-wild input due to the large-scale data it trains on, as well as the world-level knowledge that it inherits from the large multimodal model. \Cref{fig:reallife_supp} showcases our model's sewing pattern prediction from an in-the-wild image with GPT-generated text descriptions.  

\subsection{Sewing Pattern Prediction from Texts}
\begin{figure}[htp!]
    \centering
    \includegraphics[width=\textwidth]{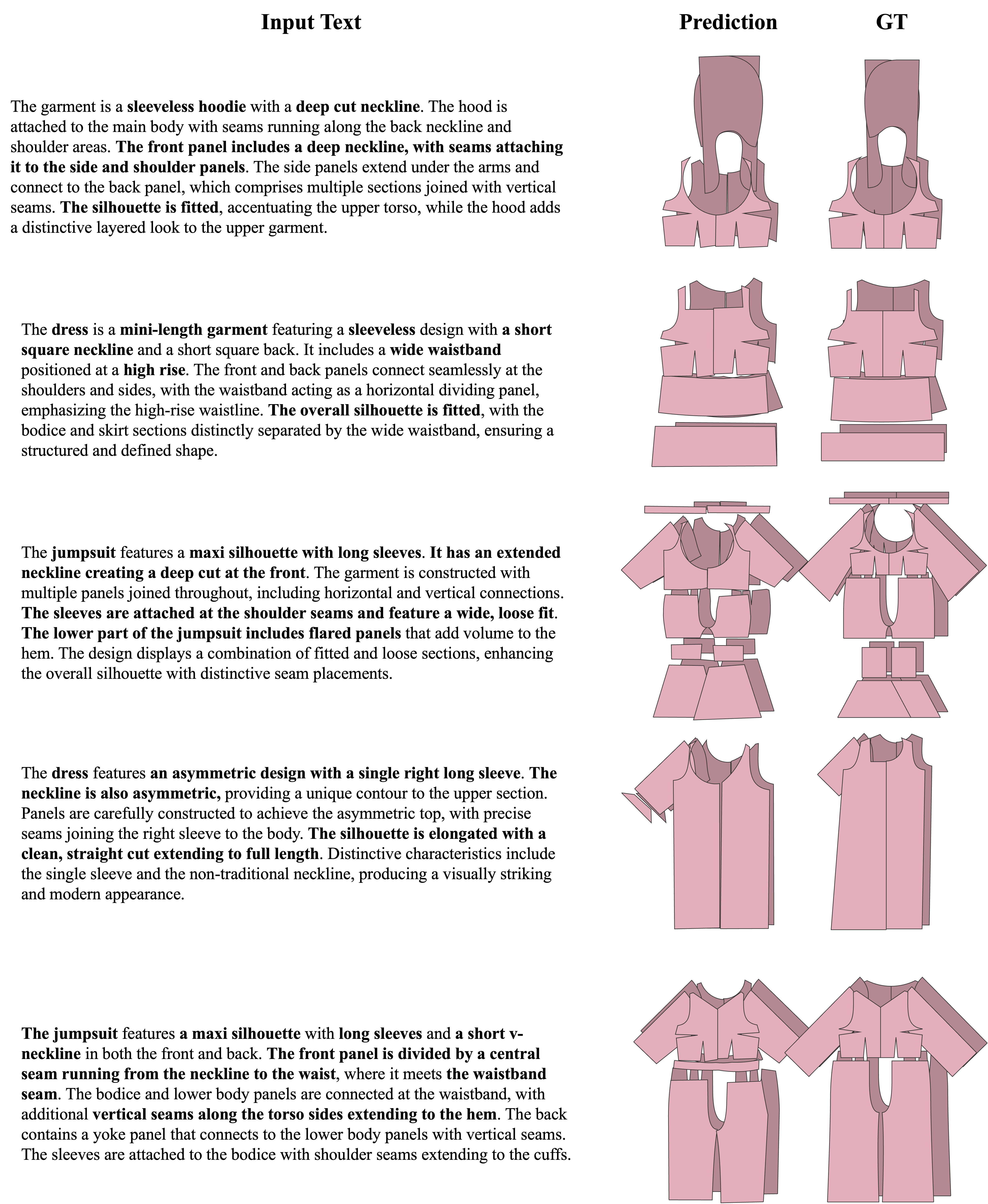}
    \caption{\textbf{Text-conditioned sewing pattern generation.} \methodname generates accurate sewing patterns closely following the text descriptions. Notice that the characteristics described in the bolded phrases all appear in the generated sewing patterns.}
    \label{fig:text2garment_supp}
\end{figure}

We showcase additional text-to-sewing pattern generation visualization from \methodname in \Cref{fig:text2garment_supp}. 
Notice that our method is able to output correct sewing patterns from long, detailed text descriptions. Moreover, our generated sewing patterns also closely follow the key characteristics described in the text input. 

\subsection{Sewing Pattern Prediction from Multimodal Input}
\paragraph{Setup.}
For our multimodal evaluation, we utilize 20 samples for each of the following modality combinations: (1) image, (2) text, (3) image + text, (4) occasion, and (5) editing. These samples are generated following the procedure outlined in \Cref{sec:dataset_detail}. To ensure proper testing, these test samples are entirely distinct from the training and validation sets used in other experiments.

To benchmark our method, we compare it against two state-of-the-art baselines: SewFormer and DressCode. SewFormer processes image-based inputs, while DressCode is designed for text-based inputs. Since these baselines are limited to specific modalities, we convert multimodal inputs into formats compatible with their architectures. For SewFormer, we use DALL-E 2 to generate a single 512x512 image from non-image inputs using tailored prompts. For DressCode, we convert inputs into keyword-based formats with GPT-4o.

The evaluation of our method and these baselines is conducted using Garment Accuracy, a metric defined as the product of Panel Accuracy and Edge Accuracy, which quantifies the percentage of garments reconstructed with the correct number of panels and edges. Additionally, we measure the squared distance between the predicted and ground-truth vertex positions to assess the geometric accuracy of the reconstructions.

\paragraph{Baselines.}
To generate an image input from a non-image modality, we use DALL-E 2 to produce a single 512x512 image. The prompt used for generation always begins with:
\begin{tcolorbox}[colback=promptbg, colframe=promptborder, boxrule=0.5mm, arc=2mm, left=2mm, right=2mm]
Create an image of a single garment worn by a mannequin. The mannequin should be front-facing and in t-pose.
\end{tcolorbox}
The prompt is tailored to each input modality by appending the following continuations.

\begin{tcolorbox}[colback=promptbg, colframe=promptborder, boxrule=0.5mm, arc=2mm, left=2mm, right=2mm]
\begin{itemize}
    \item \textbf{Text:}  Make sure that the garment follows this description: + text
    \item \textbf{Occasion:} Make sure the garment suits the following occasion: + text
    \item \textbf{Editing:} Make sure the garment looks like if this edit + edit + was applied to the garment.
\end{itemize}
\end{tcolorbox}

Similarly, to convert any input modality into a keyword-based format compatible with DressCode, we design distinct prompts based on the modality. Each prompt is constructed as a concatenation of the following starting phrase:
\begin{tcolorbox}[colback=promptbg, colframe=promptborder, boxrule=0.5mm, arc=2mm, left=2mm, right=2mm]
Describe the garment in a list of comma separated keywords. Give a maximum of 5 keywords.
\end{tcolorbox}
and a modality specific continuation:
\begin{tcolorbox}[colback=promptbg, colframe=promptborder, boxrule=0.5mm, arc=2mm, left=2mm, right=2mm]
\begin{itemize}
    \item \textbf{Text:}  Make sure that the garment follows this description: + text
    \item \textbf{Image:} Make sure that the garment looks like this image.
    \item \textbf{Text + Image:} Make sure that the garment looks like this image and follows this description: + text.
    \item \textbf{Occasion:} Make sure the garment suits the following occasion: + text
    \item \textbf{Editing:} Make sure the garment looks like if this edit + edit + was applied to the garment.
\end{itemize}
\end{tcolorbox}

\begin{figure}[htp!]
    \centering
    \includegraphics[width=\textwidth]{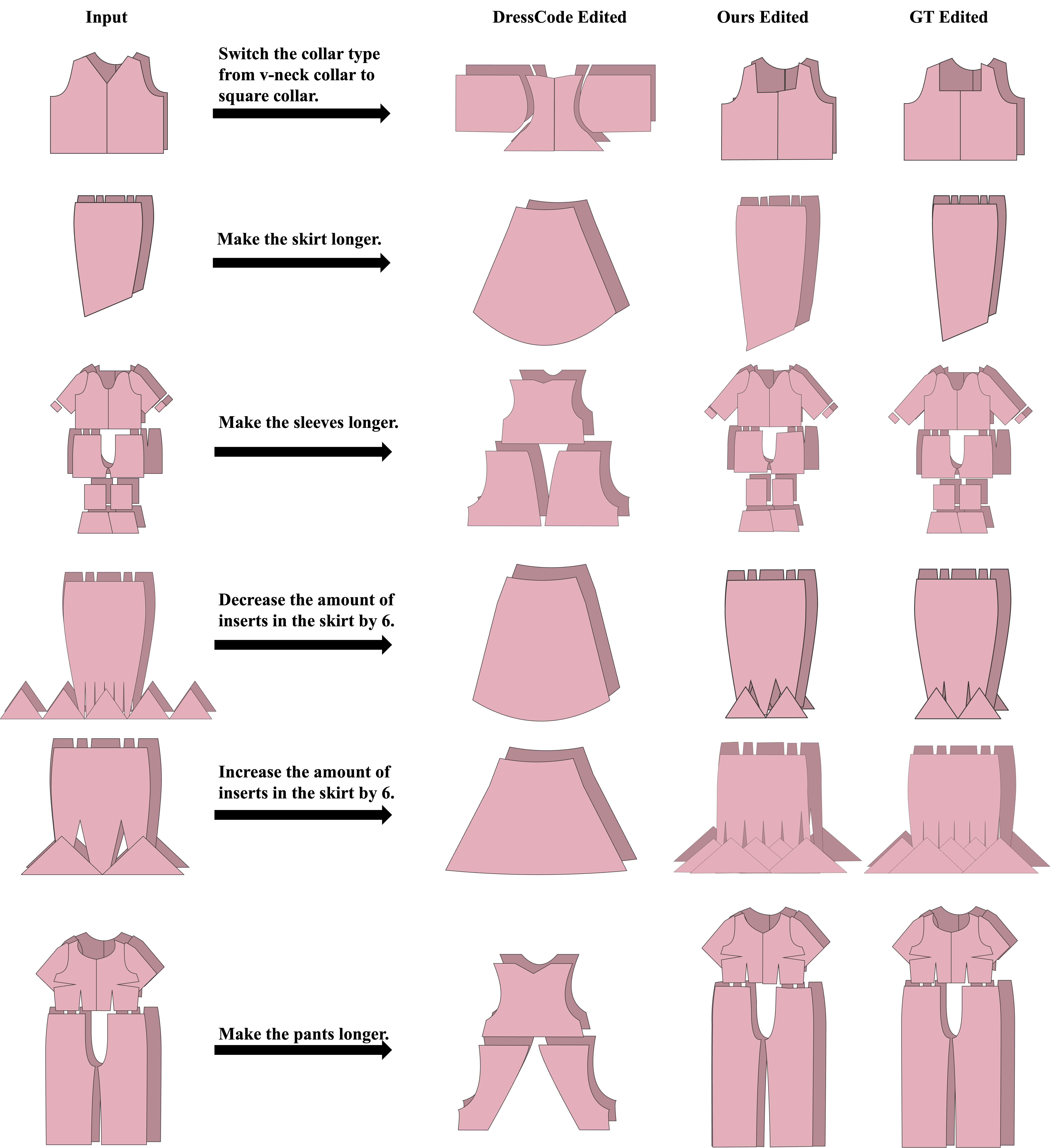}
    \caption{\textbf{Additional Visualization for Sewing Pattern Editing.} The task is to predict a sewing pattern that closely matches the input sewing pattern while following the editing instructions (text above the arrow). Notice that despite the diverse kinds of editing instructions we give, our methods can output sewing patterns that closely follow the instructions and the input sewing pattern. In the meanwhile, the baseline cannot achieve a similar effect because it takes only takes in text as input, losing structural details.}
    \label{fig:edit_supp}
\end{figure}

% \paragraph{Additional Qualitative Visualization}
% We showcase additional multimodal prediction result comparisons to the baselines. \todo{See Fig.~\ref{}}.
\subsection{Sewing Pattern Editing}
We detail the baseline methods we used for \Cref{tab:editing} and \Cref{fig:editing} in the main paper. 
Using existing models, we extend SewFormer and DressCode to translate the sewing pattern and editing instructions to their input domains. Specifically, for SewFormer, we take the editing instruction and rendered image from GCD-MM and translate the rendering image using a pre-trained InstructPix2Pix~\cite{brooks2022instructpix2pix} with the editing instruction as input. The output from InstructPix2Pix is a garment image generated based on the editing instructions and the input rendering. 
With this input image, we query the SewFormer-FT baseline to obtain the final sewing pattern. For DressCode, we use GPT4V to translate the editing instructions and rendered image into short keywords that describe the edited garment. This is then used to query the pre-trained DressCode and obtain the sewing pattern. The text prompt we use for querying GPT4V is the following: 
\begin{tcolorbox}[colback=promptbg, colframe=promptborder, boxrule=0.5mm, arc=2mm, left=2mm, right=2mm]
You are given a list of attributes describing a garment. Your task is to modify the list according to an editing instruction provided.

To accomplish this:
1. If the attribute related to the instruction already exists in the description, locate and modify it to reflect the new information.
2. If the attribute is not present, add a new entry to the description that fulfills the instruction.
3. Ensure that no other attributes are altered unless necessary for consistency or clarity following the modification.

Once the changes are complete, return the list of attributes, without any additional information.
\end{tcolorbox}

We evaluate this task using the test split of GCD-MM, containing approximately 6,000 editing samples.

\paragraph{Additional Qualitative Visualization}
\Cref{fig:edit_supp} shows additional visualization of the editing tasks as shown in \Cref{fig:editing} of the main paper. Notice that our model is able to correctly edit the sewing pattern with a diverse set of instructions. 

\subsection{Ablation Study}
\paragraph{Setup \& Baseline Details.}
\Cref{tab:ablation} in the main paper shows an ablation study on our proposed tokenization scheme in \Cref{sec:training} of the main paper. 
As described in \Cref{sec:ablation}, we use text-to-image as our ablation task to conduct an equal comparison of our model with DressCode~\cite{he2024dresscodeautoregressivelysewinggenerating}'s pre-trained model. 
Futhermore, we swap our tokenizer into DressCode's model, to ensure an equal comparison. 
We also do the same for the configuration, \textit{Ours w.o. reg.}, which uses the proposed sewing pattern tokenization scheme without the usage regression heads. 
We train both models from scratch with a learning rate of 0.0006 and a total batch size of 512 on 2$\times$Quadro RTX 8000 GPUs, for a total of 30,400 steps until convergence.  
 
\input{tables/lora_ablation}
\input{tables/layer_ablation}

\paragraph{Additional Ablation Study.}
\Cref{tab:lora_ablation} shows a qualitative comparison studying the effectiveness of full model fine-tuning versus LoRA~\cite{hu2021loralowrankadaptationlarge} fine-tuning. 
The table reports reconstruction metrics on the image-to-garment prediction task on GCD-MM dataset. For the LoRA model, we use rank 8 and only fine-tune the query and key projection layers following previous works~\cite{feng2024chatpose, lai2023lisa}. 
The model is trained with the same hyperparameter settings described in \Cref{sec:supp_training_detal} for 8250 steps. 
The metrics indicate that the full fine-tuning model significantly outperforms the LoRA fine-tuned version, indicating that fine-tuning all weights in the language transformer is essential for understanding sewing patterns. 

\begin{figure}[tp!]
    \centering
    \includegraphics[width=\textwidth]{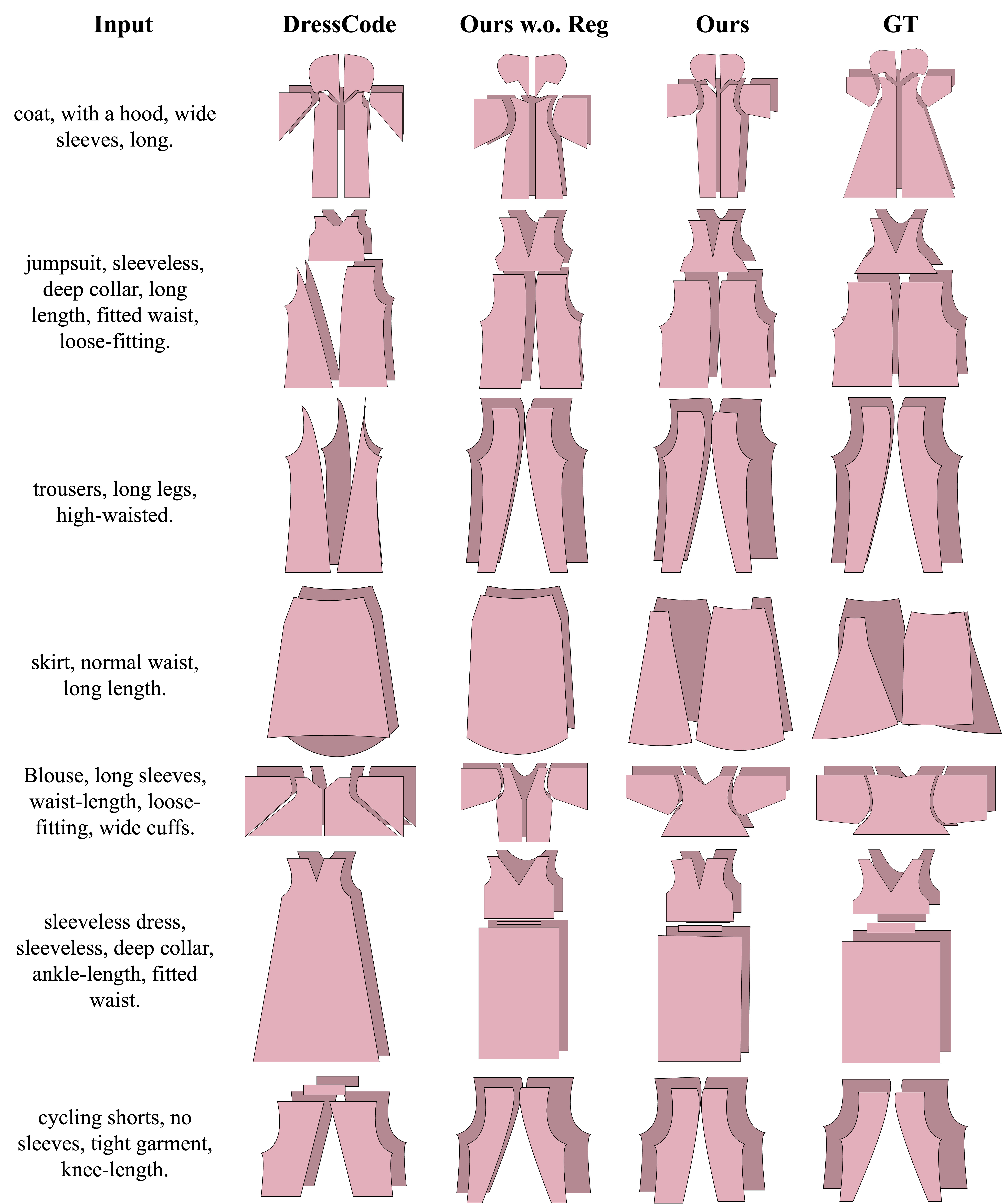}
    \caption{\textbf{Additional Visualizations for Ablation Study.}}
    \label{fig:ablation_supp}
\end{figure}

\paragraph{Additional Qualitative Visualization.}
\Cref{fig:ablation_supp} shows additional visualizations for the ablation study in \Cref{sec:ablation} of the main paper. 
Notice that our model in general demonstrates better sewing pattern prediction ability than DressCode. 
This can be seen in the pants prediction in the second and third rows of the figure, where DressCode does not predict the correct sewing pattern. 

\subsection{Draping Details}
We use the draping pipeline provided by GarmentCode~\cite{korosteleva2023garmentcode} for converting sewing patterns to a 3D mesh of the garment draped on a standard female SMPL model in A-pose. 
Specifically, the draping process consists of creating the boxed mesh and using Nvidia-Warp~\cite{warp2022} for cloth simulation. 
To obtain the garment in arbitrary poses and in a motion sequence, we follow the simulation pipeline provided by PhysAvatar~\cite{PhysAavatar24}, which uses Codimensional Incremental Potential Contact (C-IPC)~\cite{Li2021CIPC} simulation for cloth simulation. For simulation details, please refer to \citet{PhysAavatar24}. 
Finally, the simulated mesh sequence is imported to Blender for texturing and rendering. 

\subsection{Human Study}
\begin{figure}[h]
    \centering
\includegraphics[width=\linewidth]{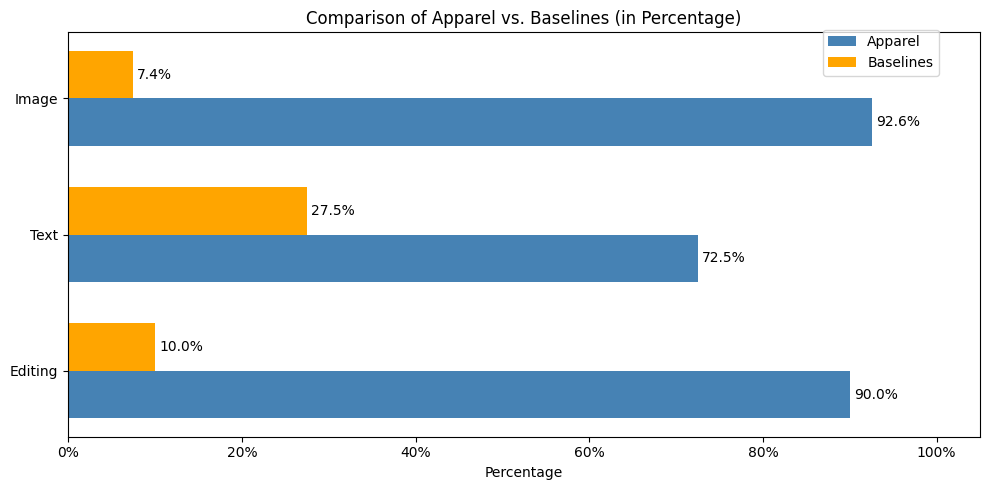}
    \caption{\textbf{User Favorability Comparison of Apparel vs. Baselines for Multimodal Generation.}}
    \label{fig:user_study}
\end{figure}
We conducted a user study to compare sewing patterns generated from multimodal inputs using AIpparel with those using baselines. Specifically, we deployed 10 multiple-choice questions asking which garment better aligns with the input prompts while maintaining realism. The questions contain a combination of sewing patterns generated from images, texts, and editions of existing sewing patterns from different methods. We collected responses from 73 participants and Fig.~\ref{fig:user_study} shows the favorability comparison for each modes of generation. AIpparel is more favorable in all modes, aligning with our quantitative and qualitative results.

\section{Discussion}
We expand our discussion in Sec.~\ref{sec:discussion} of the main paper to include further limitations, future work, and social impact.  

\subsection{More Discussion on Limitations and Future Work}
% Besides the limitations we discussed in the main paper, we here list out further future directions. 

Due to computational resource constraints, we only train \methodname on part of the GCD data, and \methodname outputs a single modality, sewing pattern.
As the community gets more computing resources, we are excited to see works extending our methods to larger datasets with richer annotations.
It is an interesting direction to further scale up \methodname to study the emergence of abilities like few-shot or in-context generalization to novel garment generation tasks or perform chain-of-thoughts to achieve a complex garment design.
It is also an interesting direction to study how to further enlarge sewing pattern datasets with more variations and more annotations.
For example, reflecting realistic variations of fabric properties can enable more accurate sewing pattern prediction.
\paragraph{Bias and Comprehensiveness of GCD-MM.}
AIpparel can inherit the bias from the sewing pattern dataset used to create GCD-MM. In fact, GarmentCodeData~\cite{korosteleva2024garmentcodedatadataset3dmadetomeasure} discusses such biases in its limitation section including only sewing patterns fitted to statistical models computed from a pool of healthy European and North American adults, hence limiting the size variations within the sewing patterns of GCD. However, we note that our data curation pipeline outlined in the paper can be used for other sources. By applying our pipeline to other, less biased, and more comprehensive sewing pattern datasets, we can still improve their quality by creating annotations for the sewing patterns.

% First, due to computational resource constraints, \methodname is only trained on the default body type of sewing patterns present in GarmentCodeData. 
% This constrains all the sewing patterns to be fitted to the same body type, limiting the model's practicality for generating garments for a larger target audience. 
% GarmentCodeData also released random body-fitted sewing patterns, and it would be a promising future direction to also include those sewing patterns as training data for \methodname to enable further generalizability of our model. 

% We limited this project to sewing pattern generation. However, apriori we can also predict other modalities as output together with sewing patterns. 
% By adding more diverse multimodal data samples, we can extend \methodname's understanding of sewing patterns and hopefully develop reasoning ability that can achieve novel-task generalization. 
% It is an interesting direction to develop a model that can output vision-language-garment signals.  

\subsection{Further Societal Impacts}
% We also expand on our discussion of social impacts. 
Besides the concerns of hallucination and bias that we inherit from our base model, LLaVA, we also acknowledge that our generated sewing patterns might not produce suitable garments for all communities, due to the limited body type and style selections within the data we trained on. 
% However, this is an inherent limitation of the sewing pattern collection provided by GarmentCodeData. 
It is important to study how to improve our method and dataset annotation on more diverse sewing patterns and body types in the future. 

Another potential risk of our work is the potential bias we inherit from foundation models in our annotation generation process. 
Because we use large models such as GPT-4V for data generation, existing biases in these models will also be included in our generated annotations. 
However, because the prompts we used (see \Cref{sec:text_detail}) encourage the model to generate descriptions based on the given images and keyword phrases, we did not find any immediate systematic bias present in our annotations. 

%% file: tables/dataset_stats.tex
\begin{table}[t]
    \centering
    \begin{tabular}{l@{}ccc}
    \toprule
          &  SewFactory~\cite{liu2023sewformer} & DressCode's Dataset~\cite{he2024dresscodeautoregressivelysewinggenerating, KorostelevaGarmentData}& \textbf{GCD-MM} \\
         \midrule
        % Panel Stats & \multicolumn{3}{c}{Fig.~\ref{fig:dataset_stats} (a)}\\
        % Edge Stats &   \multicolumn{3}{c}{Fig.~\ref{fig:dataset_stats} (b)}\\
        % Stitching Stats &  \multicolumn{3}{c}{Fig.~\ref{fig:dataset_stats} (c)}\\
        Edge Types & L, QB & L, QB & L, QB, CB, A \\
        Number of Sewing Patterns & 13700 & 20292 & 127629 \\
    \bottomrule
    \end{tabular}
    \vspace{-0.75em}
    \caption{
    \textbf{Dataset Statistics Comparison.} L=Line, QB=Quadratic Beziér, CB=Cubic Beziér, A=Arc.  GCD-MM shows a larger variation in both numbers of panels, edges, and stitches than previous sewing pattern datasets. For Panel, edge, and stitching statistics, refer to \Cref{fig:dataset_stats}.
    % We show text-to-garment metrics using the NeuralTailor dataset following DressCode's experiment setup.
    % Our tokenizer outperforms DressCode in all metrics while being more than 25 times faster at inference time.
    % Our objective (Eq.~\ref{eq:obj}) also improves performance compared to the cross-entropy-only variant. 
    % \todo{NeuralTailor data, pretrained DressCode, text-to-garment reconstruction. GPT train from scratch.} \todo{Show our data too?}
    % \vspace{-0.7cm}
    }
    \label{tab:dataset_stats}
\end{table}

%% file: tables/lora_ablation.tex
\begin{table*}[t!]
    \centering
    \begin{tabular}{l@{}cccccc@{}}
    \toprule
         Method &  Panel L2 $(\downarrow)$ & \#Panel Acc $(\uparrow)$ & \#Edge Acc $(\uparrow)$  & Rot L2 $(\downarrow)$ & Transl L2 $(\downarrow)$ & \#Stitch Acc $(\uparrow)$\\\midrule
    % SewFormer pretrained & & & & && \\    
    % DressCode pretrained (+ GPT4V)  & & & & &&\\
    % Sewformer-pretrained \\
    LoRA & 13.7 & 31.6 & 45.4 & .020 & 5.1 & .088\\
    \textbf{\methodname}   & \textbf{5.4} & \textbf{85.2} & \textbf{82.7} & \textbf{.020} & \textbf{2.7} & \textbf{77.2} \\
    \bottomrule
    \end{tabular}
    \vspace{-0.5em}
    \caption{
    \textbf{Ablation Study: Fine-tuning Comparison.}  
    % We report the reconstruction performance on the Sewfactory and the GarmentCode datasets.
    % For the SewFactory dataset~\cite{liu2023sewformer}, we compare \methodname to the pre-trained SewFormer model.
    % For the GarmentCode dataset~\cite{korosteleva2024garmentcodedatadataset3dmadetomeasure}, we compare \methodname to SewFormer finetuned on the GarmentCode dataset. 
    % We compare to SewFormer trained or finetuned at the same dataset.
    The scores are reported on the image-to-garment prediction tasks on GCD-MM dataset. The metrics indicate that full model fine-tuning significantly outperforms LoRA fine-tuning, allowing the base model to better adapt to sewing pattern understanding. 
    % We report the garment reconstruction metrics of both Sewformer and our model in \todo{GarmentCode~\cite{korosteleva2023garmentcode} dataset, which contains complex and real-world garments}. Our model achieves better performance in all metrics.\todo{On Sewformer's dataset itself?}
    }
    \label{tab:lora_ablation}
\end{table*}

%% file: tables/layer_ablation.tex
\begin{table*}[t!]
    \centering
    \begin{tabular}{l@{}cccccc@{}}
    \toprule
         Method &  Panel L2 $(\downarrow)$ & \#Panel Acc $(\uparrow)$ & \#Edge Acc $(\uparrow)$  & Rot L2 $(\downarrow)$ & Transl L2 $(\downarrow)$ & \#Stitch Acc $(\uparrow)$\\\midrule
    % SewFormer pretrained & & & & && \\    
    % DressCode pretrained (+ GPT4V)  & & & & &&\\
    % Sewformer-pretrained \\
    6 layers &  5.93 & 83.6 & 81.0 & .008 & 2.9 & 74.3\\
    5 layers & 6.10 & 84.2 & 80.7 & 0.010 & 2.8 & 73.4\\
    4 layers & 5.94 & 83.2 & 81.3 & 0.011 & 3.0 & 74.7\\
    3 layers & 5.92 & 83.7 & 80.9 & 0.010 & 2.9 & 73.7\\
    \textbf{2 layers} & \textbf{5.4} & \textbf{85.2} & \textbf{82.7} & \textbf{.020} & \textbf{2.7} & \textbf{77.2}\\
    \bottomrule
    \end{tabular}
    \vspace{-0.5em}
    \caption{
    \textbf{Ablation Study: Number of Layers in Regression Heads.}  
    % We report the reconstruction performance on the Sewfactory and the GarmentCode datasets.
    % For the SewFactory dataset~\cite{liu2023sewformer}, we compare \methodname to the pre-trained SewFormer model.
    % For the GarmentCode dataset~\cite{korosteleva2024garmentcodedatadataset3dmadetomeasure}, we compare \methodname to SewFormer finetuned on the GarmentCode dataset. 
    % We compare to SewFormer trained or finetuned at the same dataset.
    The scores are reported on the image-to-garment prediction tasks on GCD-MM dataset. 
    % The metrics indicate that full model fine-tuning significantly outperforms LoRA fine-tuning, allowing the base model to better adapt to sewing pattern understanding. 
    % We report the garment reconstruction metrics of both Sewformer and our model in \todo{GarmentCode~\cite{korosteleva2023garmentcode} dataset, which contains complex and real-world garments}. Our model achieves better performance in all metrics.\todo{On Sewformer's dataset itself?}
    }
    \label{tab:num_layer_ablation}
\end{table*}